\documentclass[format=acmsmall, review=false, screen=true]{acmart}

\usepackage{booktabs} 

\usepackage[ruled]{algorithm2e} 

\SetAlFnt{\small}
\SetAlCapFnt{\small}
\SetAlCapNameFnt{\small}
\SetAlCapHSkip{0pt}
\IncMargin{-\parindent}

\acmJournal{CSUR}
\acmVolume{0}
\acmNumber{0}
\acmArticle{0}
\copyrightyear{2018}

\setcopyright{acmcopyright}

\acmDOI{0000001.0000001}

\received{April 2018}
\received[Revised]{October 2018}
\received[Accepted]{October 2018}

\begin{document}
\title{A Comprehensive Survey of Deep Learning for Image Captioning}

\author{Md. Zakir Hossain}
\orcid{0000-0003-1212-4652}
\affiliation{%
  \institution{Murdoch University}
  \country{Australia}}
\email{MdZakir.Hossain@murdoch.edu.au}
\author{Ferdous Sohel}
\affiliation{%
  \institution{Murdoch University}
  \country{Australia}
}
\email{F.Sohel@murdoch.edu.au}
\author{Mohd Fairuz Shiratuddin}
\affiliation{%
 \institution{Murdoch University}
 \country{Australia}}
\email{f.shiratuddin@murdoch.edu.au}
\author{Hamid Laga}
\affiliation{%
  \institution{Murdoch University}
  \country{Australia}
}
\email{H.Laga@murdoch.edu.au}

\renewcommand\shortauthors{Hossain et al.}

\begin{abstract}
Generating a description of an image is called image captioning. Image captioning requires to recognize the important objects, their attributes and their relationships in an image. It also needs to generate syntactically and semantically correct sentences. Deep learning-based techniques are capable of handling the complexities and challenges of image captioning. In this survey paper, we aim to present a comprehensive review of existing deep learning-based image captioning techniques. We discuss the foundation of the techniques to analyze their performances, strengths and limitations. We also discuss the datasets and the evaluation metrics popularly used in deep learning based automatic image captioning.
\end{abstract}

%
%


%


\maketitle

\section{Introduction}

Every day, we encounter a large number of images from various sources such as the internet, news articles, document diagrams and advertisements. These sources contain images that viewers would have to interpret themselves. Most images do not have a description, but the human can largely understand them without their detailed captions. However, machine needs to interpret some form of image captions if humans need automatic image captions from it.

 Image captioning is important for many reasons. For example, they can be used for automatic image indexing. Image indexing is important for Content-Based Image Retrieval (CBIR) and therefore, it can be applied to many areas, including biomedicine, commerce, the military, education, digital libraries, and web searching. Social media platforms such as Facebook and Twitter can directly generate descriptions from images. The descriptions can include where we are (e.g., beach, cafe), what we wear and importantly what we are doing there.

Image captioning is a popular research area of Artificial Intelligence (AI) that deals with image understanding and a language description for that image. Image understanding needs to detect and recognize objects. It also needs to understand scene type or location, object properties and their interactions. Generating well-formed sentences requires both syntactic and semantic understanding of the language \cite{Vinyals17}. 
 
 Understanding an image largely depends on obtaining image features. The techniques used for this purpose can be broadly divided into two categories: (1) Traditional machine learning based techniques and (2) Deep machine learning based techniques.

 In traditional machine learning, hand crafted features such as Local Binary Patterns (LBP) \cite{Ojala00}, Scale-Invariant Feature Transform (SIFT) \cite{Lowe04}, the Histogram of Oriented Gradients (HOG) \cite{Dalal05},  and a combination of such features are widely used. In these techniques, features are extracted from input data. They are then passed to a classifier such as Support Vector Machines (SVM) \cite{Boser92} in order to classify an object. Since hand crafted features are task specific, extracting features from a large and diverse set of data is not feasible. Moreover, real world data such as images and video are complex and have different semantic interpretations.

 On the other hand, in deep machine learning based techniques, features are learned automatically from training data and they can handle a large and diverse set of images and videos. For example, Convolutional Neural Networks (CNN) \cite{LeCun98}  are widely used for feature learning, and a classifier such as Softmax is used for classification. CNN is generally followed by Recurrent Neural Networks (RNN) in order to generate captions.

In the last 5 years, a large number of articles have been published on image captioning  with deep machine learning being popularly used. Deep learning algorithms can handle complexities and challenges of image captioning quite well. So far, only three survey papers \cite{bernardi2016,kumar2017,bai2018survey} have been published on this research topic. Although the papers have presented a good literature survey of image captioning, they could only cover a few papers on deep learning because the bulk of them was published after the survey papers. These survey papers mainly discussed template based, retrieval based, and a very few deep learning-based novel image caption generating models. However, a large number of works have been done on deep learning-based image captioning. Moreover, the availability of large and new datasets has made the learning-based image captioning an interesting research area. To provide an abridged version of the  literature, we present a survey mainly focusing on the deep learning-based papers on image captioning.

 The main aim of this paper is to provide a comprehensive survey of deep learning for image captioning. First, we group the existing image captioning articles into three main categories: (1) Template-based Image captioning, (2) Retrieval-based image captioning, and (3) Novel image caption generation. The categories are discussed briefly in Section 2. Most deep learning based image captioning methods fall into the category of novel caption generation. Therefore, we focus only on novel caption generation with deep learning. Second, we group the deep learning-based image captioning methods into different categories namely (1) Visual space-based, (2) Multimodal space-based, (3) Supervised learning, (4) Other deep  learning, (5) Dense captioning, (6) Whole scene-based, (7) Encoder-Decoder Architecture-based, (8) Compositional Architecture-based, (9) LSTM (Long Short-Term Memory) \cite{Hochreiter97} language model-based, (10) Others language model-based, (11) Attention-Based, (12) Semantic concept-based, (13) Stylized captions, and (12) Novel object-based image captioning. We discuss all the categories in Section 3. We provide an overview of the datasets and commonly used evaluation metrics for measuring the quality of image captions in Section 4. We also discuss and compare the results of different methods in Section 5. Finally, we give a brief discussion and future research directions in Section 6 and then a conclusion in Section 7.

\begin{figure*}
\includegraphics[width=12cm,height=8cm,keepaspectratio]{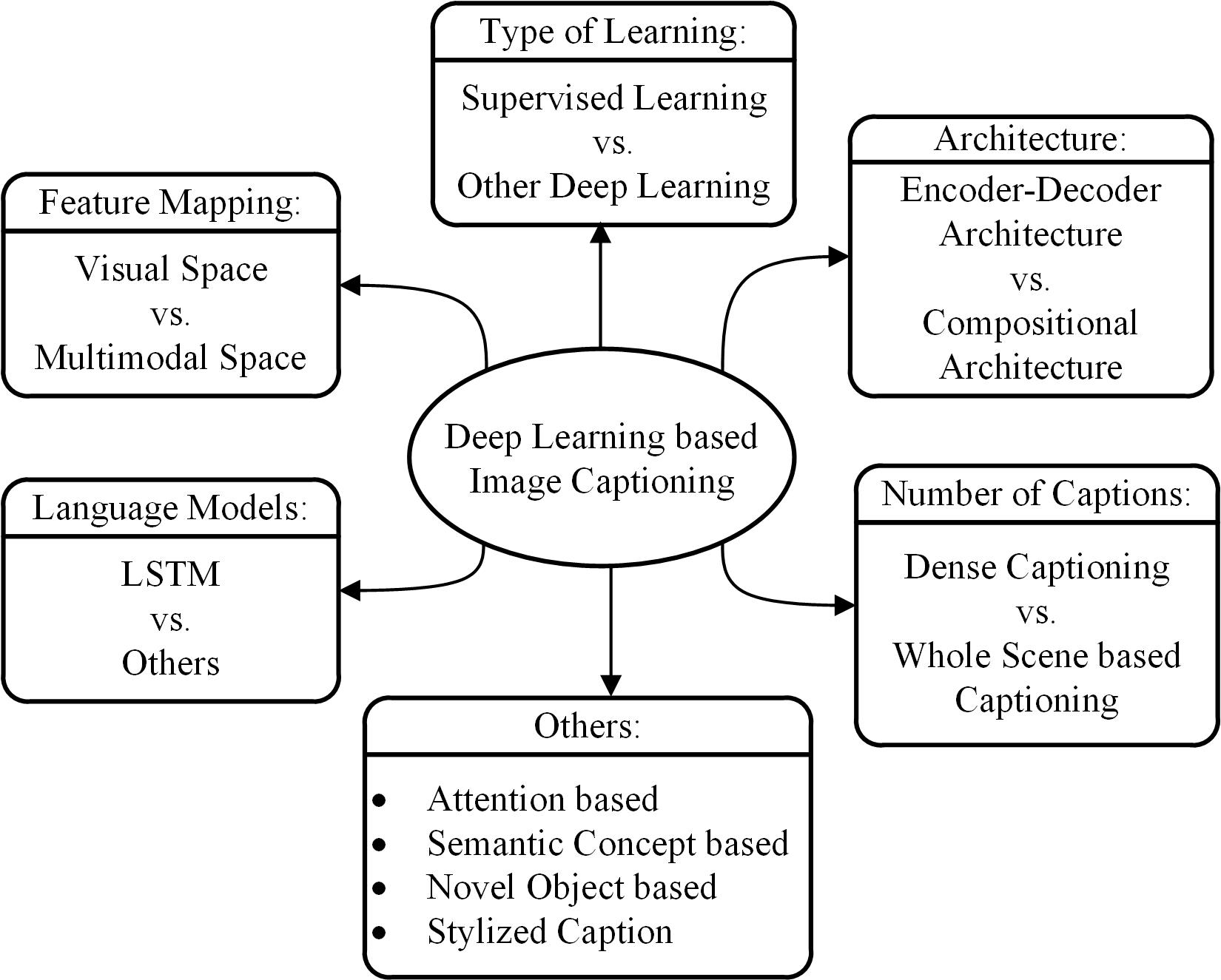}
\caption{An overall taxonomy of deep learning-based image captioning.}
\end{figure*}

\section{ Image Captioning Methods}
In this section, we review and describe the main categories of existing image captioning methods and they include template-based image captioning, retrieval-based image captioning, and novel caption generation. 
Template-based approaches have fixed templates with a number of blank slots to generate captions. In these approaches,  different objects, attributes, actions are detected first and then the blank spaces in the templates are filled. For example, Farhadi et al. \cite{Farhadi10} use a triplet of scene elements to fill the template slots for generating image captions. Li et al. \cite{Li11} extract the phrases related to detected objects, attributes and their relationships for this purpose. A Conditional Random Field (CRF) is adopted by Kulkarni et al. \cite{Kulkarni11} to infer the objects, attributes, and prepositions before filling in the gaps. Template-based methods can generate grammatically correct captions. However, templates are predefined and cannot generate variable-length captions. Moreover, later on, parsing based language models have been introduced in image captioning \cite{aker10,Elliott13,Kuznetsova12, kuznetsova14,mitchell12} which are more powerful than fixed template-based methods. Therefore, in this paper, we do not focus on these template based methods.

Captions can be retrieved from visual space and multimodal space. In retrieval-based approaches, captions are retrieved from  a set of existing captions. Retrieval based methods first find the visually similar images with their captions from the training data set. These captions are called candidate captions. The captions for the query image are selected from these captions pool \cite{ ordonez11, hodosh13,sun15,gong14}. These methods produce general and syntactically correct captions. However, they cannot generate image specific and semantically correct captions. 

Novel captions can be generated from both visual space and multimodal space. A general approach of this category is to analyze the visual content of the image first and then generate image captions from the visual content using a language model \cite{Kiros14U,Xu15,Yao16,You16}. These methods can generate new captions for each image that are semantically more accurate than previous approaches. Most novel caption generation methods use deep machine learning based techniques. Therefore, deep learning based novel image caption generating methods are our main focus in this literature.

An overall taxonomy of deep learning-based image captioning methods is depicted in Figure 1. The figure illustrates the comparisons of different categories of image captioning methods. Novel caption generation-based image caption methods mostly use visual space and deep machine learning based techniques. Captions can also be generated from multimodal space. Deep learning-based image captioning methods can also be categorized on learning techniques: Supervised learning, Reinforcement learning, and Unsupervised learning. We group the reinforcement learning and unsupervised learning into Other Deep Learning. Usually captions are generated for a whole scene in the image. However, captions can also be generated for different regions of an image (Dense captioning). Image captioning methods can use either simple Encoder-Decoder architecture or Compositional architecture. There are methods that use attention mechanism, semantic concept, and different styles in image descriptions. Some methods can also generate description for unseen objects. We group them into one category as ``Others". Most of the image captioning methods use LSTM as language model. However, there are a number of methods that use other language models such as CNN and RNN. Therefore, we include a language model-based category as ``LSTM vs. Others".

\section{Deep Learning Based Image Captioning Methods}

We draw an overall taxonomy in Figure 1 for deep learning-based image captioning methods. We discuss their similarities and dissimilarities by grouping them into visual space vs. multimodal space, dense captioning vs. captions for the whole scene, Supervised learning vs. Other deep learning, Encoder-Decoder architecture vs. Compositional architecture, and one `Others' group that contains Attention-Based, Semantic Concept-Based, Stylized captions, and Novel Object-Based captioning. We also create a category named LSTM vs. Others.

A brief overview of the deep learning-based image captioning methods is shown in Table 1. Table 1 contains the name of the image captioning methods, the type of deep neural networks used to encode image information, and the language models used in describing the information. In the final column, we give a category label to each captioning technique based on the taxonomy in Figure 1.

\subsection{Visual Space vs. Multimodal Space }

Deep learning-based image captioning methods can generate captions from both visual space and multimodal space. Understandably image captioning datasets have the corresponding captions as text. In the visual space-based methods, the image features and the corresponding captions are independently passed to the language decoder. In contrast, in a multimodal space case, a shared multimodal space is learned from the images and the corresponding caption-text. This multimodal representation is then passed to the language decoder. 

\subsubsection{Visual Space}
Bulk of the image captioning methods use visual space for generating captions. These methods are discussed in Section 3.2 to Section 3.5.

\begin{figure*}
\includegraphics[width=12cm,height=8cm,keepaspectratio]{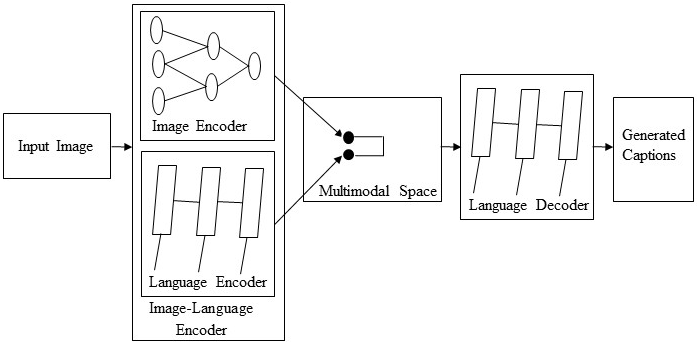}
\caption{A block diagram of multimodal space-based image captioning.}
\end{figure*}

\subsubsection{Multimodal Space}
The architecture of a typical multimodal space-based method contains a language Encoder part, a vision part, a multimodal space part, and a language decoder part. A general diagram of multimodal space-based image captioning methods is shown in Figure 2. The vision part uses a deep convolutional neural network as a feature extractor to extract the image features. The language encoder part extracts the word features and learns a dense feature embedding for each word. It then forwards the semantic temporal context to the recurrent layers. The multimodal space part maps the image features into a common space with the word features. The resulting map is then passed to the language decoder which generates captions by decoding the map.
 
 The methods in this category follow the following steps:
\begin{enumerate}
\item Deep neural networks and multimodal neural language model are used to learn both image and text jointly in a multimodal space.
\item  The language generation part generates captions using the information from Step 1 .
\end{enumerate}

\begin{table*}[]
\centering
\renewcommand{\arraystretch}{0.8}
\begin{tabular}{|l|l|l|l|l|}
\hline
\multirow{2}{*}{\textbf{Reference}}                      & \multirow{2}{*}{\textbf{Image Encoder}} & \multirow{2}{*}{\textbf{\begin{tabular}[c]{@{}l@{}}Language Model\end{tabular}}} & \multicolumn{2}{l|}{\multirow{2}{*}{\textbf{Category}}} \\
                                                         &                                         &                                                                                     & \multicolumn{2}{l|}{}                 \\ \hline
Kiros et al. 2014~\cite{Kiros14M} & AlexNet                        & LBL                                                         & \multicolumn{2}{l|}{MS,SL,WS,EDA}          \\ \hline
Kiros et al. 2014~\cite{Kiros14U}  & AlexNet, VGGNet                & \begin{tabular}[c]{@{}l@{}}1. LSTM\\ 2. SC-NLM\end{tabular} & \multicolumn{2}{l|}{MS,SL,WS,EDA}          \\ \hline
Mao et al. 2014~\cite{Mao14E}    & AlexNet                        & RNN                                                         & \multicolumn{2}{l|}{MS,SL,WS}          \\ \hline
Karpathy et al. 2014~\cite{Karpathy14}  & AlexNet                        & DTR                                                         & \multicolumn{2}{l|}{MS,SL,WS,EDA}          \\ \hline
Mao et al. 2015~\cite{Mao14}         & AlexNet, VGGNet                & RNN                                                         & \multicolumn{2}{l|}{MS,SL,WS}          \\ \hline
Chen et al. 2015~\cite{Chen15}   & VGGNet                         & RNN                                                         & \multicolumn{2}{l|}{VS,SL,WS,EDA}                 \\ \hline
Fang et al. 2015~\cite{Fang15} & AlexNet, VGGNet                & MELM                                                        & \multicolumn{2}{l|}{VS,SL,WS,CA}                 \\ \hline
Jia et al. 2015~\cite{jia2015}    & VGGNet                         & LSTM                                                        & \multicolumn{2}{l|}{VS,SL,WS,EDA}                  \\ \hline
Karpathy et al. 2015~\cite{Karpathy15}  & VGGNet                         & RNN                                                         & \multicolumn{2}{l|}{MS,SL,WS,EDA}                  \\ \hline
Vinyals et al. 2015~\cite{Vinyals15}     & GoogLeNet                      & LSTM                                                        & \multicolumn{2}{l|}{VS,SL,WS,EDA}                \\ \hline
Xu et al. 2015~\cite{Xu15}          & AlexNet                        & LSTM                                                        & \multicolumn{2}{l|}{VS,SL,WS,EDA,AB}                 \\ \hline
Jin et al. 2015~\cite{Jin15}     & VGGNet                         & LSTM                                                        & \multicolumn{2}{l|}{VS,SL,WS,EDA,AB}                 \\ \hline
Wu et al. 2016~\cite{Wu16E}  & VGGNet                         & LSTM                                                        & \multicolumn{2}{l|}{VS,SL,WS,EDA,AB}                 \\ \hline
Sugano et at. 2016~\cite{sugano16}& VGGNet                         & LSTM                                                        & \multicolumn{2}{l|}{VS,SL,WS,EDA,AB}                 \\ \hline
Mathews et al. 2016~\cite{mathews16} & GoogLeNet                      & LSTM                                                        & \multicolumn{2}{l|}{VS,SL,WS,EDA,SC}                  \\ \hline
Wang et al. 2016~\cite{wang16}   & AlexNet, VGGNet                & LSTM                                                        & \multicolumn{2}{l|}{VS,SL,WS,EDA}                \\ \hline
Johnson et al. 2016~\cite{ johnson16} & VGGNet                         & LSTM                                                        & \multicolumn{2}{l|}{VS,SL,DC,EDA}          \\ \hline
Mao et al. 2016~\cite{ mao16}   & VGGNet                         & LSTM                                                        & \multicolumn{2}{l|}{VS,SL,WS,EDA}                \\ \hline
Wang et al. 2016 \cite{wang2016parallel} & VGGNet                & LSTM                                                    &  \multicolumn{2}{l|}{VS,SL,WS,CA}          \\ \hline
Tran et al. 2016~\cite{tran16}  & ResNet                         & MELM                                                        & \multicolumn{2}{l|}{VS,SL,WS,CA}                 \\ \hline
Ma et al. 2016 \cite{ma2016describing}  & AlexNet                         & LSTM                                                        & \multicolumn{2}{l|}{VS,SL,WS,CA}                 \\ \hline
You et al. 2016~\cite{You16}        & GoogLeNet                      & RNN                                                         & \multicolumn{2}{l|}{VS,SL,WS,EDA,SCB} \\ \hline
Yang et al. 2016~\cite{yang16}       & VGGNet                         & LSTM                                                        & \multicolumn{2}{l|}{VS,SL,DC,EDA}          \\ \hline
Anne et al. 2016~\cite{anne2016deep}        & VGGNet                     & LSTM                                                         & \multicolumn{2}{l|}{VS,SL,WS,CA,NOB}                  \\ \hline
Yao et al. 2017~\cite{Yao16}  & GoogLeNet                      & LSTM                                                        & \multicolumn{2}{l|}{VS,SL,WS,EDA,SCB}                  \\ \hline
Lu et al. 2017~\cite{Lu16}   & ResNet                         & LSTM                                                        & \multicolumn{2}{l|}{VS,SL,WS,EDA,AB}                 \\ \hline
Chen et al. 2017~\cite{Chen16}        & VGGNet, ResNet                 & LSTM                                                        & \multicolumn{2}{l|}{VS,SL,WS,EDA,AB}                 \\ \hline
Gan et al. 2017~\cite{gan16sem}     & ResNet                         & LSTM                                                        & \multicolumn{2}{l|}{VS,SL,WS,CA,SCB}                 \\ \hline
Pedersoli et al. 2017~\cite{ Pedersoli16}  & VGGNet                         & RNN                                                         & \multicolumn{2}{l|}{VS,SL,WS,EDA,AB}                 \\ \hline
Ren et al. 2017~\cite{ren17}         & VGGNet                         & LSTM                                                        & \multicolumn{2}{l|}{VS,ODL,WS,EDA}    \\ \hline
Park et al. 2017~\cite{Park17}      & ResNet                         & LSTM                                                        & \multicolumn{2}{l|}{VS,SL,WS,EDA,AB}                 \\ \hline
Wang et al. 2017~\cite{ wang17}    & ResNet                         & LSTM                                                        & \multicolumn{2}{l|}{VS,SL,WS,EDA}                 \\ \hline
Tavakoli et al. 2017~\cite{ Tavakoli17}  & VGGNet                         & LSTM                                                        & \multicolumn{2}{l|}{VS,SL,WS,EDA,AB}                 \\ \hline
Liu et al. 2017~\cite{ Liu17}    & VGGNet                         & LSTM                                                        & \multicolumn{2}{l|}{VS,SL,WS,EDA,AB}                 \\ \hline
Gan et al. 2017~\cite{gan17}     & ResNet                         & LSTM                                                        & \multicolumn{2}{l|}{VS,SL,WS,EDA,SC}         \\ \hline
Dai et al. 2017~\cite{ dai17}      & VGGNet                         & LSTM                                                        & \multicolumn{2}{l|}{VS,ODL,WS,EDA}                       \\ \hline
Shetty et al. 2017~\cite{ shetty17}  & GoogLeNet                      & LSTM                                                        & \multicolumn{2}{l|}{VS,ODL,WS,EDA}                       \\ \hline
Liu et al. 2017~\cite{liu2017improved}    & Inception-V3                         & LSTM                                                        & \multicolumn{2}{l|}{VS,ODL,WS,EDA}                 \\ \hline
Gu et al. 2017~\cite{gu2017empirical} & VGGNet                & \begin{tabular}[c]{@{}l@{}}1. Language CNN\\ 2. LSTM\end{tabular} & \multicolumn{2}{l|}{VS,SL,WS,EDA}          \\ \hline
Yao et al. 2017~\cite{yao2017incorporating}    & VGGNet                         & LSTM                                                        & \multicolumn{2}{l|}{VS,SL,WS,CA,NOB}                 \\ \hline
Rennie et al. 2017~\cite{rennie2017self}    & ResNet                         & LSTM                                                        & \multicolumn{2}{l|}{VS,ODL,WS,EDA}                 \\ \hline
Vsub et al. 2017~\cite{venugopalan2017captioning}    & VGGNet                         & LSTM                                            & \multicolumn{2}{l|}{VS,SL,WS,CA,NOB}                 \\ \hline
Zhang et al. 2017 ~\cite{zhang2017actor}    & Inception-V3                         & LSTM                                            & \multicolumn{2}{l|}{VS,ODL,WS,EDA}                 \\ \hline
Wu et al. 2018~\cite{ wu2018image}         & VGGNet                         & LSTM                                                        & \multicolumn{2}{l|}{VS,SL,WS,EDA,SCB}                  \\ \hline
Aneja et al. 2018 \cite{aneja2018convolutional}    & VGGNet                         & Language CNN                                            & \multicolumn{2}{l|}{VS,SL,WS,EDA}                 \\ \hline
Wang et al. 2018 \cite{wang2018cnn+}    & VGGNet                         & Language CNN                                            & \multicolumn{2}{l|}{VS,SL,WS,EDA}                 \\ \hline

\end{tabular}
\bigskip
\caption{An overview of the deep learning-based approaches for image captioning (VS=Visual Space, MS=Multimodal Space, SL=Supervised Learning, ODL=Other Deep Learning, DC=Dense Captioning, WS=Whole Scene, EDA=Encoder-Decoder Architecture, CA=Compositional Architecture, AB=Attention-Based, SCB=Semantic Concept-Based, NOB=Novel Object-Based, SC=Stylized Caption).}
\label{Table1}
\end{table*}

An initial work in this area proposed by Kiros et al. \cite{Kiros14M}. The method applies a CNN for extracting image features in generating image captions. It uses a multimodal space that represents both image and text jointly for multimodal representation learning and image caption generation. It also introduces the multimodal neural language models such as Modality-Biased Log-Bilinear Model (MLBL-B) and the Factored 3-way Log-Bilinear Model (MLBL-F) of \cite{Mnih07} followed by AlexNet \cite{Krizhevsky12}. Unlike most previous approaches, this method does not rely on any additional templates, structures, or constraints. Instead it depends on the high level image features and word representations learned from deep neural networks and multimodal neural language models respectively. The neural language models have limitations to handle a large amount of data and are inefficient to work with long term memory \cite{Jozefowicz16}. 

Kiros et al. \cite{Kiros14M} extended their work in \cite{Kiros14U} to learn a joint image sentence embedding where LSTM is used for sentence encoding and a new neural language model called the structure-content neural language model (SC-NLM) is used for image captions generations. The SC-NLM has an advantage over existing methods in that it can extricate the structure of the sentence to its content produced by the encoder. It also helps them to achieve significant improvements in generating realistic image captions over the approach proposed by \cite{Kiros14M}

Karpathy et al. \cite{Karpathy14} proposed a deep, multimodal model, embedding of image and natural language data for the task of bidirectional images and sentences retrieval. The previous multimodal-based methods use a common, embedding space that directly maps images and sentences. However, this method works at a finer level and embeds fragments of images and fragments of sentences. This method breaks down the images into a number of objects and sentences into a dependency tree relations (DTR) \cite{DeMarneffe06} and reasons about their latent, inter-modal alignment. It shows that the method achieves significant improvements in the retrieval task compared to other previous methods. This method has a few limitations as well. In terms of modelling, the dependency tree can model relations easily but they are not always appropriate. For example, a single visual entity might be described by a single complex phrase that can be split into multiple sentence fragments. The phrase ``black and white dog'' can be formed into two relations (CONJ, black, white) and (AMOD, white, dog). Again, for many dependency relations we do not find any clear mapping in the image (For example: ``each other'' cannot be mapped to any object).

Mao et al. \cite{Mao14} proposed a multimodal Recurrent Neural Network (m-RNN) method for generating novel image captions. This method has two sub-networks: a deep recurrent neural network for sentences and a deep convolutional network for images. These two sub-networks interact with each other in a multimodal layer to form the whole m-RNN model. Both image and fragments of sentences are given as input in this method. It calculates the probabilty distribution to generate the next word of captions. There are five more layers in this model: Two-word embedding layers, a recurrent layer, a multimodal layer and a SoftMax layer. Kiros et al. \cite{Kiros14M} proposed a method that is built on a Log-Bilinear model and used AlexNet to extract visual features. This multimodal recurrent neural network method is closely related to the method of Kiros et al. \cite{Kiros14M}. Kiros et al. use a fixed length context (i.e. five words), whereas in this method, the temporal context is stored in a recurrent architecture, which allows an arbitrary context length. The two word embedding layers use one hot vector to generate a dense word representation. It encodes both the syntactic and semantic meaning of the words. The semantically relevant words can be found by calculating the Euclidean distance between two dense word vectors in embedding layers. Most sentence-image multimodal methods \cite{Karpathy14,Frome13,Socher14, Kiros14U} use pre-computed word embedding vectors to initialize their model. In contrast, this method randomly initializes word embedding layers and learn them from the training data. This helps them to generate better image captions than the previous methods. Many image captioning methods \cite{Mao14E,Kiros14M,Karpathy14} are built on recurrent neural networks at the contemporary times. They use a recurrent layer for storing visual information. However, (m-RNN) use both image representations and sentence fragments to generate captions. It utilizes the capacity of the recurrent layer more efficiently that helps to achieve a better performance using a relatively small dimensional recurrent layer.

Chen et al. \cite{Chen15} proposed another multimodal space-based image captioning method. The method can generate novel captions from image and restore visual features from the given description. It also can describe a bidirectional mapping between images and their captions. Many of the existing methods \cite{hodosh13,Socher14,Karpathy14} use joint embedding to generate image captions. However, they do not use reverse projection that can generate visual features from captions. On the other hand, this method dynamically updates the visual representations of the image from the generated words. It has an additional recurrent visual hidden layer with RNN that makes reverse projection.

\subsection{Supervised Learning vs. Other Deep Learning }

In supervised learning, training data come with desired output called \textit{label}. Unsupervised learning, on the other hand, deals with unlabeled data. Generative Adversarial Networks (GANs) \cite{goodfellow14} are a type of unsupervised learning techniques. Reinforcement learning is another type of machine learning approach where the aims of an agent are to discover data and/or labels through exploration and a reward signal. A number of image captioning methods use reinforcement learning and GAN based approaches. These methods sit in the category of ``Other Deep  Learning". 

\subsubsection{Supervised Learning-Based Image Captioning}

Supervised learning-based networks have successfully been used for many years in image classification \cite{Krizhevsky12,he16,Simonyan14,Szegedy15}, object detection \cite{girshick15,girshick14,Ren15}, and attribute learning \cite{gan16}. This progress makes researchers interested in using them in automatic image captioning \cite{Vinyals15,Mao14,Karpathy15,Chen15}. In this paper, we have identified a large number of supervised learning-based image captioning methods. We classify them into different categories:  (i) Encoder-Decoder Architecture, (ii) Compositional Architecture, (iii) Attention-based, (iv) Semantic concept-based, (v) Stylized captions, (vi) Novel object-based, and (vii) Dense image captioning.

\begin{figure*}
\includegraphics[width=1 \linewidth]{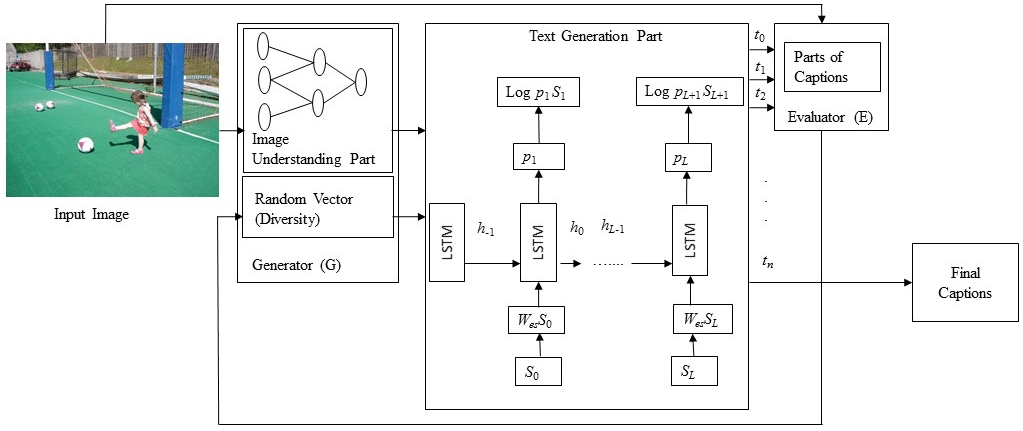}
\caption{A block diagram of other deep learning-based captioning.}
\end{figure*}

\subsubsection{Other Deep Learning-Based Image Captioning}

In our day to day life, data are increasing with unlabled data because it is often impractical to accurately annotate data. Therefore, recently, researchers are focusing more on reinforcement learning and unsupervised learning-based techniques for image captioning.

A reinforcement learning agent chooses an action, receives reward values, and moves to a new state. The agent attempts to select the action with the expectation of having a maximum long-term reward. It needs continuous state and action information, to provide the guarantees of a value function.  Traditional reinforcement learning approaches face a number of limitations such as the lack of guarantees of a value function and uncertain state-action information. Policy gradient methods \cite{sutton00} are a type of reinforcement learning that can choose a specific policy for a specific action using gradient descent and optimization techniques. The policy can incorporate domain knowledge for the action that guarantees convergence. Thus, policy gradient methods require fewer parameters than value-function based approaches. \\ Existing deep learning-based image captioning methods use variants of  image encoders to extract image features. The features are then fed into the neural network-based language decoders to generate captions. The methods have two main issues: (i) They are trained using maximum likelihood estimation and back-propagation~\cite{ranzato2016sequence} approaches. In this case, the next word is predicted given the image and all the previously generated ground-truth words. Therefore, the generated captions look-like ground-truth captions. This phenomenon is called exposure bias~\cite{bengio2015scheduled} problem. (ii) Evaluation metrics at test time are non-differentiable. Ideally sequence models for image captioning should be trained to avoid exposure bias and directly optimise metrics for the test time. In actor-critic-based reinforcement learning algorithm, critic can be used in estimating the expected future reward to train the actor (captioning policy network). Reinforcement learning-based image captioning methods sample the next token from the model based on the rewards they receive in each state. Policy gradient methods in reinforcement learning can optimize the gradient in  order to predict the cumulative long-term rewards. Therefore, it can solve the non-differentiable problem of evaluation metrics.
 
The methods in this category follow the following steps:
\begin{enumerate}
\item  A CNN and RNN based combined network generates captions.
\item Another CNN-RNN based network evaluates the captions and send feedback to the first network to generate high quality captions.
\end{enumerate}

A block diagram of a typical method of this category is shown in Figure 3.

Ren et al. 2017 \cite{ren17} introduced a novel reinforcement  learning based image captioning method. The architecture of this method has two networks that jointly compute the next best word at each time step. The ``policy network'' works as local guidance and helps to predict next word based on the current state. The ``value network''' works as global guidance and evaluates the reward value considering all the possible extensions of the current state. This mechanism is able to adjust the networks in predicting the correct  words. Therefore, it can generate good captions similar to ground truth captions at the end. It uses an actor-critic reinforcement learning model \cite{konda00} to train the whole network. Visual semantic embedding \cite{ ren15multi,ren16} is used to compute the actual reward value in predicting the correct word. It also helps to measure the similarity between images and sentences that can evaluate the correctness of generated captions.

Rennie et al. \cite{rennie2017self} proposed another reinforcement learning based image captioning method. The method utilizes the test-time inference algorithm to normalize the reward rather than estimating the reward signal and normalization in training time. It shows that this test-time decoding is highly effective for generating quality image captions.

Zhang et al. \cite{zhang2017actor} proposed an actor-critic reinforcement learning-based image captioning method. The method can directly optimize non-differentiable problems of the existing evaluation metrics. The architecture of the actor-critic method consists of a policy network (actor) and a value network (critic). The actor treats the job as sequential decision problem and can predict the next token of the sequence. In each state of the sequence, the network will receive a task-specific reward (in this case, it is evaluation metrics score). The job of the critic is to predict the reward. If it can predict the expected reward, the actor will continue to sample outputs according to its probability distribution.

GAN based methods can learn deep features from unlabeled data. They achieve this representations applying a competitive process between a pair of networks: the Generator and the Discriminator. GANs have already been used successfully in a variety of applications, including image captioning\cite{dai17,shetty17}, image to image translation \cite{isola16}, text to image synthesis \cite{reed16, bodnar2018text}, and text generation \cite{fedus2018maskgan,wang2018text}.

There are two issues with GAN. First, GAN can work well in generating natural images from  real  images because GANs are proposed for real-valued data. However, text processing is based on discrete numbers. Therefore, such operations are non-differentiable, making it difficult to apply back-propagation directly. Policy gradients apply a parametric function to allow gradients to be back-propagated. Second, the evaluator faces problems in vanishing gradients and error propagation for sequence generation. It needs a probable future reward value for every partial description. Monte Carlo rollouts~\cite{yu16} is used to compute this future reward value.
  
 GAN based image captioning methods can generate a diverse set of image captions in contrast to conventional deep convolutional network and deep recurrent network based model. Dai et al. \cite{dai17} also  proposed a GAN based image captioning method. However, they do not consider multiple captions for a single image. Shetty et al. \cite{shetty17} introduced a new GAN based image captioning method. This method can generate multiple captions for a single image and showed  impressive improvements in  generating diverse captions. GANs have limitations in backpropagating the discrete data. Gumbel sampler \cite{jang16,maddison16} is used to overcome the discrete data problem. The two main parts of this adversarial network are the generator and the discriminator. During training, generator learns the loss value provided by the discriminator instead of learning it from explicit sources. Discriminator has true data distribution and can discriminate between generator-generated samples and true data samples. This allows the network to learn diverse data distribution. Moreover, the network classifies the generated caption sets either real or fake. Thus, it can generate captions similar to human generated one.

\begin{figure*}
\includegraphics[width=1 \linewidth]{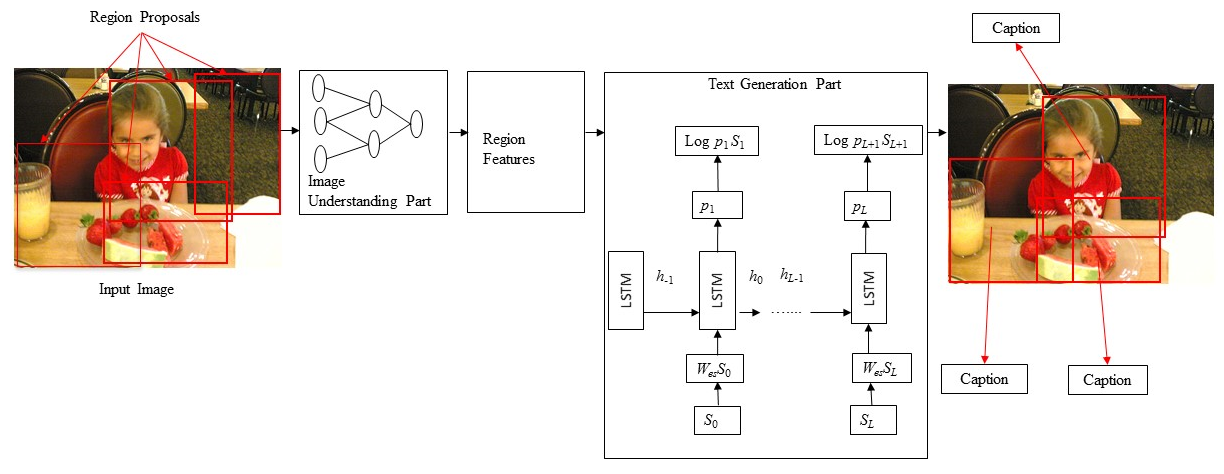}
\caption{A block diagram of dense captioning.}
\end{figure*}
\subsection{Dense Captioning vs. Captions for the whole scene }
 In dense captioning, captions are generated for each region of the scene. Other methods generate captions for the whole scene.
\subsubsection{Dense Captioning}
 The previous image captioning methods can generate only one caption for the whole image. They use different regions of the image to obtain information of various objects. However, these methods do not generate region wise captions. 
 
 Johnson et al. \cite {johnson16} proposed an image captioning method called DenseCap. This method localizes all the salient regions of an image and then it generates descriptions for those regions. 

  A typical method of this category has the following steps:
\begin{enumerate}
\item  Region proposals are generated for the different regions of the given image.
\item  CNN is used to obtain the region-based image features.
\item  The outputs of Step 2 are used by a language model to generate captions for every region.
\end{enumerate}

 A block diagram of a typical dense captioning method is given in Figure 4.
 
 Dense captioning~\cite{johnson16} proposes a fully convolutional localization network architecture, which is composed of a convolutional network, a dense localization layer, and an LSTM \cite{Hochreiter97} language  model. The dense localization layer processes an image with a single, efficient forward pass, which implicitly predicts a set of region of interest in the image. Thereby, it requires no external region proposals unlike to Fast R-CNN or a full network (i.e., RPN (Region Proposal Network \cite{girshick15})) of Faster R-CNN. The working principle of the  localization layer is related to the work of Faster R-CNN \cite{Ren15}. However, Johnson et al. \cite {johnson16} use a differential, spatial soft attention mechanism \cite {gregor15,jaderberg15}  and bilinear interpolation \cite {jaderberg15} instead of ROI pooling mechanism \cite{girshick15}. This modification helps the method to backpropagate through the network and smoothly select the active regions. It uses Visual Genome~\cite{krishna17} dataset for the experiments in generating region level image captions.

One description of the entire visual scene is quite subjective and is not enough to bring out the complete understanding. Region-based descriptions are more objective and detailed than global image description. The region-based description is known as dense captioning. There are some challenges in dense captioning. As regions are dense, one object may have multiple overlapping regions of interest. Moreover, it is very difficult to recognize each target region for all the visual concepts. Yang et al.  \cite{yang16} proposed another dense captioning method. This method can tackle these challenges. First, it addresses an inference mechanism that jointly depends on the visual features of the region and the predicted captions for that region. This allows the model to find an appropriate position of the bounding box. Second, they apply a context fusion that can combine context features with the visual features of respective regions to provide a rich semantic description.

\subsubsection{Captions for the whole scene}
 Encoder-Decoder architecture, Compositional architecture, attention-based, semantic concept-based, stylized captions, Novel object-based image captioning, and other deep learning networks-based image captioning methods generate single or multiple captions for the whole scene.

\subsection{Encoder-Decoder Architecture vs. Compositional Architecture}

Some methods use just simple vanilla encoder and decoder to generate captions. However, other methods use multiple networks for it.

\subsubsection{Encoder-Decoder Architecture-Based Image captioning}
 The neural network-based image captioning methods work as just simple end to end manner. These methods are very similar to the encoder-decoder framework-based neural machine translation \cite{Sutskever14}. In this network, global image features are extracted from the hidden activations of CNN and then fed them into an LSTM to generate a sequence of words.

\begin{figure*}
\includegraphics[width=12cm,height=8cm,keepaspectratio]{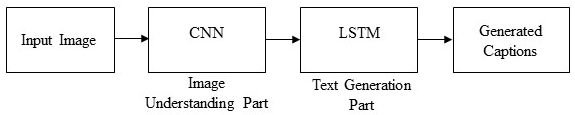}
\caption{A block diagram of simple Encoder-Decoder architecture-based image captioning.}
\end{figure*}

  A typical method of this category has the following general steps:
\begin{enumerate}
\item  A vanilla CNN is used to obtain the scene type, to detect the objects and their relationships.
\item  The output of Step 1 is used by a language model to convert them into words, combined phrases that produce an image captions.
\end{enumerate}
 A simple block diagram of this category is given in Figure 5.
 
 Vinyals  et al. \cite{Vinyals15} proposed a method called Neural Image Caption Generator (NIC). The method uses a CNN for image representations and an LSTM for generating image captions. This special CNN uses a novel method for batch normalization and the output of the last hidden layer of CNN is used as an input to the LSTM decoder. This LSTM is capable of keeping track of the objects that already have been described using text. NIC is trained based on maximum likelihood estimation.
 
 In generating image captions, image information is included to the initial state of an LSTM. The next words are generated  based on the current time step and the previous hidden state. This process continues until it gets the end token of the sentence. Since image information is fed only at the beginning of the process, it may face vanishing gradient problems. The role of the words generated at the beginning is also becoming weaker and weaker. Therefore, LSTM is still facing challenges in generating long length sentences \cite{bahdanau2014,cho2014}. Therefore, Jia et al. \cite{jia2015} proposed an extension of LSTM called guided LSTM (gLSTM). This gLSTM can generate long sentences. In this architecture, it adds global semantic information to  each gate and cell state of LSTM. It also considers different length normalization strategies to control the length of captions. Semantic information is extracted in different ways. First, it uses a cross-modal retrieval task for retrieving image captions and then semantic information is extracted from these captions. The semantic based information can also be extracted using a multimodal embedding space.

 Mao et al. \cite {mao16} proposed a special type of text generation method for images. This method can generate a description for an specific object or region that is called referring expression \cite{van06,viethen08,mitchell10,mitchell13,fitzgerald13,golland10,kazemzadeh14}. Using this expression it can then infer the object or region which is being described. Therefore, generated description or expression is quite unambiguous. In order to address the referring expression, this method uses a new dataset called ReferIt dataset \cite{kazemzadeh14} based on popular MS COCO dataset.

Previous CNN-RNN based image captioning methods  use LSTM that are unidirectional and relatively shallow in depth. In unidirectional language generation techniques, the next word is predicted based on visual context and all the previous textual contexts. Unidirectional LSTM cannot generate contextually well formed captions. Moreover, recent object detection and classification methods \cite{Krizhevsky12,Simonyan14} show that deep, hierarchical methods are better at learning than shallower ones. Wang et al. \cite{wang16} proposed a deep bidirectional LSTM-based method for image captioning. This method is capable of generating contextually and semantically rich image captions. The proposed architecture consists of a CNN and two separate LSTM networks. It can utilize both past and future context  information  to learn long term visual language interactions.

\subsubsection{Compositional Architecture-Based Image captioning}

 Compositional architecture-based methods composed of several independent functional building blocks: First, a CNN is used to extract the semantic concepts from the image. Then a language model is used to generate  a set of candidate captions. In generating the final caption, these candidate captions are re-ranked using a deep multimodal  similarity model.

\begin{figure*}
\includegraphics[width=14cm,height=10cm,keepaspectratio]{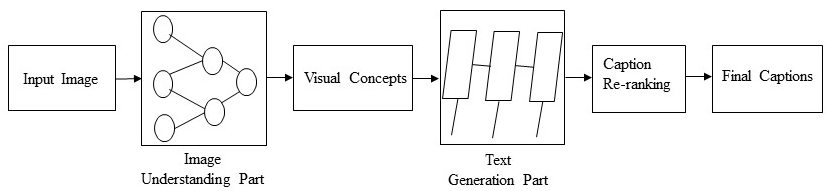}
\caption{A block diagram of a compositional network-based captioning.}
\end{figure*}
 
 A typical method of this category maintains the following steps:
\begin{enumerate}
\item  Image features are obtained using a CNN.
\item  Visual concepts (e.g. attributes) are obtained from visual features.
\item  Multiple captions are generated by a language model using the information of Step 1 and Step 2.
\item  The generated captions are re-ranked using a deep multimodal similarity model to select high quality image captions.
\end{enumerate}

A common block diagram of compositional network-based image captioning methods is given in Figure 6.

Fang et al.\cite{Fang15}  introduced generation-based image captioning. It uses visual detectors, a language model, and a multimodal similarity model to train the model on an image captioning dataset. Image captions can contain nouns, verbs, and adjectives. A vocabulary is formed using 1000 most common words from the training captions. The system works with the image sub-regions rather that the full image. Convolutional neural networks (both AlexNet \cite{Krizhevsky12} and VGG16Net) are used for extracting features for the sub-regions of an image. The features of sub-regions are mapped with the words of the vocabulary that likely to be contained in the image captions. Multiple instance learning (MIL) \cite{Maron98} is used to train the model for learning discriminative visual signatures of each word. A maximum entropy (ME) \cite{Berger96} language model is used for generating image captions from these words. Generated captions are ranked by a linear weighting of sentence features. Minimum Error rate training (MERT) \cite{Och03} is used to learn these weights. Similarity between image and sentence can be easily measured using a common vector representation. Image and sentence fragments are mapped with the common vector representation by a deep multimodal similarity model (DMSM). It achieves a significant improvement in choosing high quality image captions.

Until now a significant number of methods have achieved satisfactory progress in generating image captions. The methods use training and testing samples from the same domain. Therefore, there is no certainty that these methods can perform well in open-domain images. Moreover, they are only good at recognizing generic visual content. There are certain key entities such as celebrities and landmarks that are out of their scope. The generated captions of these methods are evaluated on automatic metrics such as BLEU \cite{papineni02}, METEOR \cite{agarwal08}, and CIDEr \cite{vedantam15}. These evaluation metrics have already shown good results on these methods. However, in terms of performance there exists a large gap between the evaluation of the metrics and human judgement of evaluation \cite{devlin2015l,callison06,Kulkarni11}. If it is considered real life entity information, the performance could be weaker. However, Tran et al. \cite{tran16} introduced a  different image captioning method. This method is capable of generating image captions even for open domain images. It can detect a diverse set of visual concepts and generate captions for celebrities and landmarks. It uses an external knowledge base Freebase  \cite{bollacker08} in recognizing a broad range of entities such as celebrities and landmarks. A series of human judgments are applied for evaluating the performances of generated captions. In experiments, it uses  three datasets: MS COCO, Adobe-MIT FiveK \cite{bychkovsky11}, and images from Instagram. The images of MS COCO dataset were collected from the same domain but the images of other datasets were chosen from an open domain. The method achieves notable performances especially on the challenging Instagram dataset.

Ma et al. \cite{ma2016describing} proposed another compositional network-based image captioning method. This method uses structural words $<$object, attribute, activity, scene$>$ to generate semantically meaningful descriptions. It also uses a multi-task method similar to multiple instance learning method \cite {Fang15}, and multi-layer optimization method \cite{han2015describing} to generate structural words. An LSTM encoder-decoder-based machine translation method \cite {Sutskever14} is then used to translate the structural words into image captions.

Wang et al. \cite{wang2016parallel} proposed a parallel-fusion RNN-LSTM architecture for image caption generation. The architecture of the method divides the hidden units of RNN and LSTM into a number of same-size parts. The parts work in parallel with corresponding ratios to generate image captions.

\subsection{Others}

Attention-based, Semantic concept-based, Novel object-based methods, and Stylized captions are put together into ``Others" group because these categories are independent to other methods.

\subsubsection{Attention based Image Captioning}  

Neural encoder-decoder based approaches were mainly used in machine translation \cite{Sutskever14}. Following these trends, they have also been used for the task of image captioning and found very effective. In image captioning, a CNN is used as an encoder to extract the visual features from the input image and an RNN is used as a decoder to convert this representation word-by-word into natural language description of the image. However, these methods are unable to analyze the image over time while they generate the descriptions for the image. In addition to this, the methods do not consider the spatial aspects of the image that is relevant to the parts of the image captions. Instead, they generate captions considering the scene as a whole. Attention based mechanisms are becoming increasingly popular in deep learning because they can address these limitations. They can dynamically focus on the various parts of the input image while the output sequences are  being produced.

\begin{figure*}
\includegraphics[width=1 \linewidth]{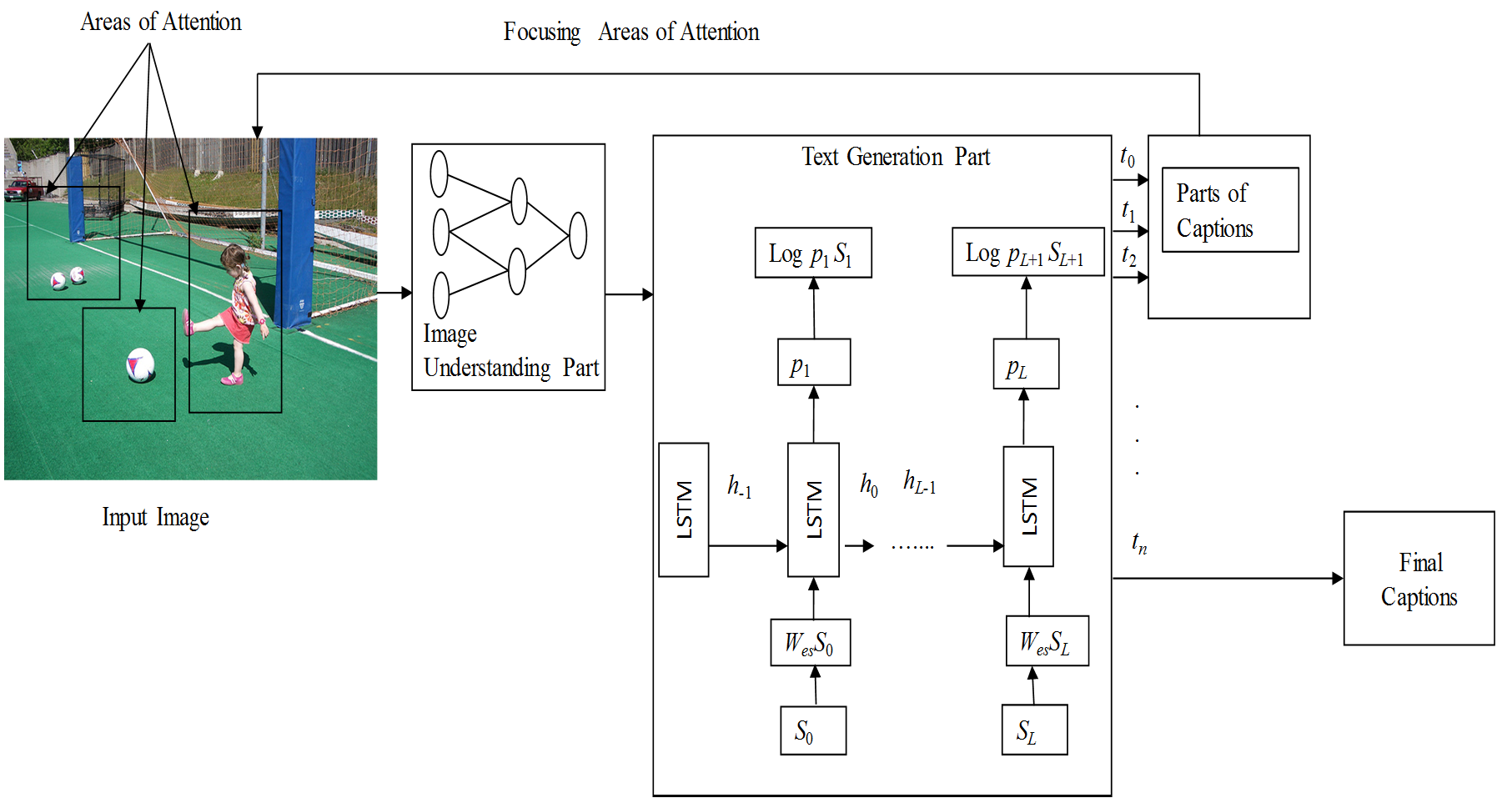}
\caption{A block diagram of a typical attention-based image captioning technique.}
\end{figure*}

A typical method of this category adopts the following steps:

\begin{enumerate}
\item  Image information is obtained based on the whole scene by a CNN.
\item  The language generation phase generates words or phrases based on the output of Step 1.
\item  Salient regions of the given image are focused in each time step of the language generation model based on generated words or phrases.
\item  Captions are updated dynamically until the end state of language generation model.

\end{enumerate}

 A block diagram of the attention-based image captioning method is shown in Figure 7.

Xu et al. \cite{Xu15} were the first to introduce an attention-based image captioning method. The method describes the salient contents of an image automatically. The main difference between the attention-based methods with other methods is that they can concentrate on the salient parts of the image and generate the corresponding words at the same time. This method applies two different techniques: stochastic hard attention and deterministic soft attention to generate attentions. Most CNN-based approaches use the top layer of ConvNet for extracting information of the salient objects from the image. A drawback of these techniques is that they may lose certain information which is useful to generate detailed captions. In order to preserve the information, the attention method uses features from the lower convolutional layer instead of fully connected layer.

Jin et al. \cite{Jin15} proposed another attention-based image captioning method. This method is capable to extract the flow of abstract meaning based on the semantic relationship between visual information and textual information. It can also obtain higher level semantic information by proposing a scene specific context. The main difference between this method with other attention-based methods is that it introduces multiple visual regions of an image at multiple scales. This technique can extract proper visual information of a particular object. For extracting scene specific context, it first uses the Latent Dirichlet Allocation (LDA) \cite{Blei03} for generating a dictionary from all the captions of the dataset. Then a multilayer perceptron is used to predict a topic vector for every image. A scene factored LSTM that has two stacked layers are used to generate a description for the overall context of the image.

Wu et al. \cite{Wu16E} proposed a review-based attention  method for image captioning. It introduces a review model that can perform multiple review steps with attention on CNN hidden states. The output of the CNN is a number of fact vectors that can obtain the global facts of the image. The vectors are given as input to the attention mechanism of the LSTM. For example, a reviewer module can first review: What are the objects in the image? Then it can review the relative positions of the objects and another review can extract the information of the overall context of the image. These information is passed to the decoder to generate image captions.

Pedersoli et al. \cite{Pedersoli16} proposed an area based attention mechanism for image captioning. Previous attention based methods map image regions only to the state of RNN language model. However, this approach associates image regions with caption words given the RNN state. It can predict the next caption word and corresponding image region in each time-step of RNN. It is capable of predicting the next word as well as corresponding image regions in each time-step of RNN for generating image captions. In order to find the areas of attention, previous attention-based image caption methods use either the position of CNN activation grid or object proposals. In contrast, this method uses an end to end trainable convolutional spatial transformer along with CNN activation gird and object proposal methods. A combination of these techniques help this method to compute image adaptive areas of attention. In experiments, the method shows that this new attention mechanism together with the spatial transformer network can produce high quality image captions.

Lu et al. \cite{Lu16}  proposed another attention-based image captioning method. The method is based on adaptive attention model with a visual sentinel. Current attention-based image captioning methods focus on the image in every time step of RNN. However, there are some words or phrase (for example: a, of) that do not need to attend visual signals. Moreover, these unnecessary visual signals could affect the caption generation process and degrade the overall performance. Therefore, their proposed method can determine when it will focus on image region and when it will just focus on language generation model. Once it determines to look on the image then it must have to choose the spatial location of the image. The first contribution of this method is to introduce a novel spatial attention method that can compute spatial features from the image. Then in their adaptive attention method, they introduced a new LSTM extension. Generally, an LSTM works as a decoder that can produce a hidden state at every time step. However, this extension is capable of producing an additional visual sentinel that provides a fallback option to the decoder. It also has a sentinel gate that can control how much information the decoder will get from the image. 

While attention-based methods look to find the different areas of the image at the time of generating words or phrases for image captions, the attention maps generated by these methods cannot always correspond to the proper region of the image. It can affect the performance of image caption generation. Liu et al. \cite{Liu17} proposed a method for neural image captioning. This method can evaluate and correct the attention map at time step. Correctness means to make consistent map between image regions and generated words. In order to achieve these goals, this method introduced a quantitative evaluation metric to compute the attention maps. It uses Flickr30k entity dataset \cite{Plummer15} and MS COCO \cite{Lin14} dataset for measuring both ground truth attention map and semantic labelings of image regions. In order to learn a better attention function, it proposed supervised attention model. Two types of supervised attention models are used here: strong supervision with alignment annotation and weak supervision with semantic labelling. In strong supervision with alignment annotation model, it can directly map ground truth word to a region. However, ground truth alignment is not always possible because collecting and annotating data is often very expensive. Weak supervision is performed to use bounding box or segmentation masks on MS COCO dataset. In experiments, the method shows that supervised attention model performs better in mapping attention as well as image captioning. 

Chen et al. \cite{Chen16} proposed another attention-based image captioning method. This method considers both spatial and channel wise attentions to compute an attention map. The existing attention-based image captioning methods only consider spatial information for generating an attention map. A common drawback of these spatial attention methods are that they compute weighted pooling only on attentive feature map. As a result, these methods lose the spatial information gradually. Moreover, they use the spatial information only from the last conv-layer of the CNN. The receptive field regions of this layer are quite large that make the limited gap between the regions. Therefore, they do not get significant spatial attentions for an image. However, in this method, CNN features are extracted not only from spatial locations but also from different channels and multiple layers. Therefore, it gets significant spatial attention. In addition to this, in this method, each filter of a convolutional layer acts as semantic detectors  \cite{Zeiler14} while other methods use external sources for obtaining semantic information. 


In order to reduce the gap between human generated description and machine generated description Tavakoli et al. \cite{Tavakoli17} introduced an attention-based image captioning method. This is a bottom up saliency based attention model that can take advantages for comparisons with other attention-based image captioning methods. It found that humans first describe the more important objects than less important ones. It also shows that the method performs better on unseen data.

 Most previous image captioning methods applied top-down approach for constructing a visual attention map. These mechanisms typically focused on some selective regions obtained from the output of one or two layers of a CNN. The input regions are of the same size and have the same shape of receptive field. This approach has a little consideration to the content of the image. However, the method of Anderson et al. \cite{Anderson17} applied both top down and bottom up approaches. The bottom up attention mechanism uses Faster R-CNN \cite{Ren15} for region  proposals that can select salient regions of an image . Therefore, this method can attend both object level regions as well as other salient image regions. 
 
Park et al. \cite{Park17}  introduced a different type of attention-based image captioning method. This method can generate image captions addressing personal issues of an image. It mainly considers two tasks : hashtag prediction and post generation. This method uses a  Context Sequence Memory Network (CSMN) to obtain the context information from the image. Description of an image from personalized view has a lot of applications in social media networks. For example, everyday people share a lot of images as posts in Facebook, Instagram or other social media. Photo-taking or uploading is a very easy task. However, describing them is not easy because it requires theme, sentiment, and context of the image. Therefore, the method considers the past knowledge about the user\textquotesingle s vocabularies or writing styles from the prior documents for generating image descriptions. In order to work with this new type of image captioning, the CSMN method has three contributions: first, the memory of this network can work as a repository and retain multiple types of context information. Second, the memory is designed in such a way that it can store all the previously generated words sequentially. As a result, it does not suffer from vanishing gradient problem. Third, the proposed CNN can correlate with multiple memory slots that is helpful for understanding contextual concepts.

Attention-based methods have already shown good performance and efficiency in image captioning as well as other computer vision tasks. However, attention maps generated by these attention based methods are only machine dependent. They do not consider any supervision from human attention. This creates the necessity to think about the gaze information whether it can improve the performance of these attention methods in image captioning. Gaze indicates the cognition and perception of humans about a scene. Human gaze can identify the important locations of objects in an image. Thus, gaze mechanisms have already shown their potential performances in eye-based user modeling \cite{ bulling11, fathi12, papadopoulos14, sattar15,shanmuga15}, object localization \cite{mishra12} or recognition \cite{ karthikeyan13} and holistic scene understanding \cite{yun13,zelinsky13}. However, Sugano et al. \cite{sugano16} claimed that gaze information has not yet been integrated in image captioning methods. This method introduced human gaze with the attention mechanism of deep neural networks in generating image captions. The method incorporates human gaze information into an attention-based LSTM model \cite{Xu15}. For experiments, it uses SALICON dataset \cite{jiang15} and achieves good results.

\begin{figure*}
\includegraphics[width=12cm,height=8cm,keepaspectratio]{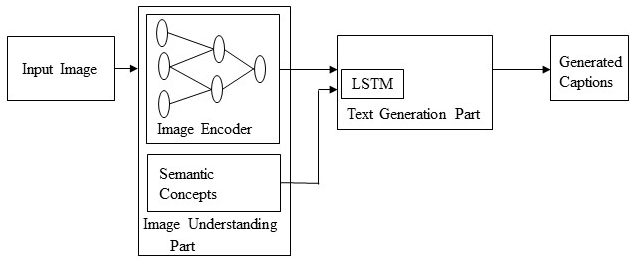}
\caption{A block diagram of a semantic concept-based image captioning.}
\end{figure*}

\subsubsection{Semantic Concept-Based Image Captioning} 

Semantic concept-based methods selectively attend to a set of semantic concept proposals extracted from the image.  These concepts are then combined into hidden states and the outputs of recurrent neural networks.

The methods in this category follow the following steps:

\begin{enumerate}
\item  CNN based encoder is used to encode the image features and semantic concepts.
\item  Image features are fed into the input of language generation model.
\item  Semantic concepts are added to the different hidden states of the language model.
\item  The language generation part produces captions with semantic concepts.

\end{enumerate}

 A typical block diagram of this category is shown in Figure 8.

Karpathy et al. extended their method \cite{Karpathy14} in \cite{Karpathy15}. The later method can generate natural language descriptions for both images as well as for their regions. This method employs a novel combination of CNN over the image regions, bidirectional Recurrent Neural Networks over sentences, and a common multimodal embedding that associates the two modalities. It also demonstrates a multimodal recurrent neural network architecture that utilizes the resultant alignments to train the model for generating novel descriptions of image regions. In this method, dependency tree relations (DTR) are used to train to map the sentence segments with the image regions that have a fixed window context. In contrast to their previous method, this method uses a bidirectional neural network to obtain word representations in the sentence. It considers contiguous fragments of sentences to align in embedding space which is more meaningful, interpretable, and not fixed in length. Generally an RNN considers the current word and the contexts from all the previously generated words for estimating a probability distribution of the next word in a sequence. However, this method extends it for considering the generative process on the content of an input image. This addition is simple but it makes it very effective for generating novel image captions.

Attributes of an image are considered as rich semantic cues. The method of Yao et al. \cite{Yao16} has different architectures to incorporate attributes with image representations. Mainly, two types of architectural representations are introduced here. In the first group, it inserts only attributes to the LSTM or image representations to the LSTM first and then attributes and vice versa. In the second group, it can control the time step of LSTM. It decides whether image representation and attributes will be inputted once or every time step. These variants of architectures are tested on MS COCO dataset and common evaluation metrics.

You et al. \cite{You16}  proposed a semantic attention-based image captioning method. The method provides a detailed, coherent description of semantically important objects. The top-down paradigms  \cite{Chen15,Vinyals15,Mao14,Karpathy15,Donahue15,Xu15,Mao15} are used for extracting visual features first and then convert them into words. In bottom up approaches,   \cite{Farhadi10,Kulkarni11,Li11,Elliott13,Kuznetsova12,Lebret14} visual concepts (e.g., regions, objects, and attributes) are extracted first from various aspects of an image and then combine them. Fine details of an image are often very important for generating a description of an image. Top- down approaches have limitations in obtaining fine details of the image. Bottom up approaches are capable of operating on any image resolution and therefore they can do work on fine details of the image. However, they have problems in formulating an end to end process. Therefore, semantic based attention model applied both top-down and bottom up approaches for generating image captions. In top-down approaches, the image features are obtained using the last 1024-dimensional convolutional layer of the GoogleNet \cite{ Szegedy15} CNN model. The visual concepts are collected using different non-parametric and parametric method. Nearest neighbour image retrieval technique is used for computing non-parametric visual concepts. Fully convolutional network (FCN) \cite{ Long15} is used to learn attribute from local patches for parametric attribute prediction. Although Xu et al. \cite{ Xu15} considered attention-based captioning, it works on fixed and pre-defined spatial location. However, this semantic attention-based method can work on any resolution and any location of the image. Moreover, this method also considers a feedback process that accelerates to generate better image captions.

 Previous image captioning methods do not include high level semantic concepts explicitly. However, Wu et al. \cite{Wu16W} proposed a high-level semantic concept-based image captioning. It uses an intermediate attribute prediction layer in a neural network-based CNN-LSTM framework. First, attributes are extracted by a CNN-based classifier from training image captions. Then these attributes are used as high level semantic concepts in generating semantically rich image captions.

\begin{figure*}
\includegraphics[width=8cm,height=4cm,keepaspectratio]{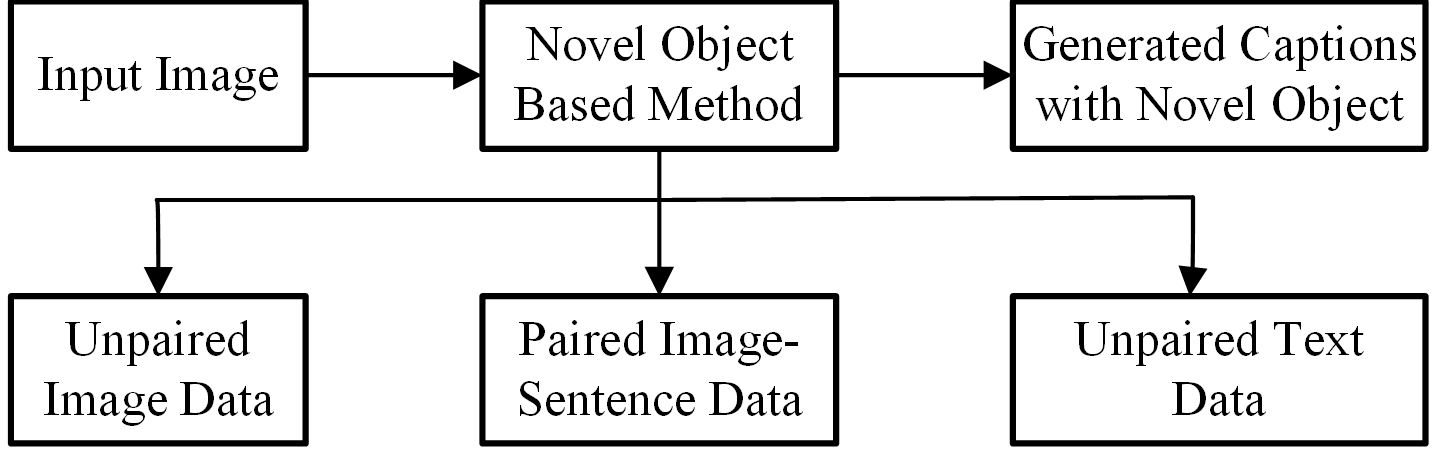}
\caption{A block diagram of a typical novel object-based image captioning.}
\end{figure*}

Recent semantic concept based image captioning methods \cite{wu2018image,You16} applied semantic-concept-detection process  \cite{gan16}to obtain explicit semantic concepts. They use these high level semantic concepts in CNN-LSTM based encoder-decoder and achieves significant improvements in image captioning. However, they have problems in generating  semantically sound captions. They cannot distribute semantic concepts evenly in the whole sentence. For example, Wu et al. \cite{wu2018image} consider the initial state of the LSTM to add semantic concepts.  Moreover, it encodes visual features vector or an inferred scene vector from the CNN and then feeds them to LSTM for generating captions. However, Gan et al. \cite{gan16sem} introduced a Semantic Compositional Network (SCN) for image captioning. In this method, a semantic concept vector is constructed from all the probable concepts (called tags here) found in the image. This semantic vector has more potential than visual feature vector and scene vector and can generate captions covering the overall meaning of the image. This is called compositional network because it can compose most semantic concepts. 

 Existing LSTM based image captioning methods have limitations in generating a diverse set of captions because they have to predict the next word on a predefined word by word format. However, a combination  of attributes, subjects and their relationship in a sentence irrespective of their location can generate a broad range of image captions. Wang et al. \cite{wang17} proposed a method that locates the objects and their interactions first and then identifies and extracts the relevant attributes to generate image captions. The main aim of this method is to decompose the ground truth image captions into two parts: Skeleton sentence and attribute phrases. The method is also called Skeleton Key. The architecture of this method has ResNet \cite{he16} and two LSTMs called Skel-LSTM and Attr-LSTM. During training, skeleton sentences are trained by Skel-LSTM network and attribute phrases are trained by the Attr-LSTM network. In the testing phase, skeleton sentences are generated first that  contain the words for main objects of the image and their relationships. Then these objects look back through the image again to obtain the relevant attributes. It is tested on MS COCO dataset and a new Stock3M dataset and can generate more accurate and novel captions.

 \subsubsection{Novel Object-based Image Captioning}
 
 Despite recent deep learning-based image captioning methods have achieved promising results, they largely depend on the paired image and sentence caption datasets. These type of methods can only generate description of the objects within the context. Therefore, the methods require a large set of training image-sentence pairs. Novel object-based image captioning methods can generate descriptions of novel objects which are not present in paired image-captions datasets.
 
 The methods of this category follow the following general steps:
\begin{enumerate}
\item  A separate lexical classifier and a language model are trained on unpaired image data and unpaired text data.
\item  A deep caption model is trained on paired image caption data.
\item  Finally, both models are combined together to train jointly in that can generate captions for novel object.
\end{enumerate}
 
 \begin{figure*}
\includegraphics[width=12cm,height=8cm,keepaspectratio]{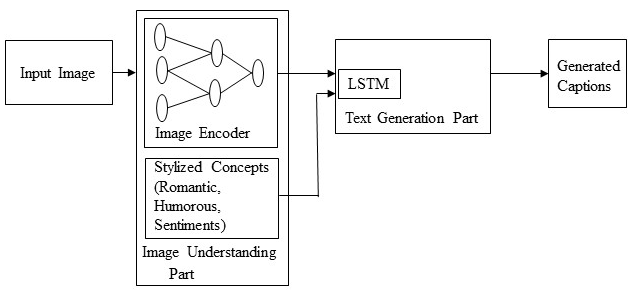}
\caption{A block diagram of image captioning based on different styles.}
\end{figure*}
 
 A simple block diagram of a novel object-based image captioning method is given in Figure 9.

 Current image captioning methods are trained on image-captions paired datasets. As a result, if they get unseen objects in the test images, they cannot present them in their generated captions. Anne et al. \cite{anne2016deep} proposed a Deep Compositional Captioner (DCC) that can represent the unseen objects in generated captions.  
 
 Yao et al. \cite{yao2017incorporating} proposed a copying mechanism to generate description for novel objects. This method uses a separate object recognition dataset to develop classifiers for novel objects. It integrates the appropriate words in the output captions by a decoder RNN with copying mechanism. The architecture of the method adds a new network to recognize the unseen objects from unpaired images and incorporate them with LSTM to generate captions.

Generating captions for the unseen images is a challenging research problem. Venugopalan et al. \cite{venugopalan2017captioning} introduced a Novel Object Captioner (NOC) for generating captions for unseen objects in the image. They used external sources for recognizing unseen objects and learning semantic knowledge.

\subsubsection{Stylized Caption}  

Existing image captioning systems generate captions just based on only the image content that can also be called factual description. They do not consider the stylized part of the text separately from other linguistic patterns. However, the stylized captions can be more expressive and attractive than just only the flat description of an image. 
 
 The methods of this category follow the following general steps:
\begin{enumerate}
\item  CNN based image encoder is used to obtain the image information.
\item  A separate text corpus is prepared to extract various stylized concepts (For example: romantic, humorous) from training data.
\item  The language generation part can generate stylized and attractive captions using the information of Step 1 and Step 2.
\end{enumerate}

A simple block diagram of stylized image captioning is given in Figure 10.

 Such captions have become popular because they are particularly valuable for many real-world applications. For example, everyday people are uploading a lot of photos in different social media. The photos need stylized and attractive descriptions. Gan et al. \cite{ gan17} proposed a novel image captioning system called StyleNet. This method can generate attractive captions adding various styles. The architecture of this method consists of a CNN and a factored LSTM that can separate factual and style factors from the captions. It uses multitask sequence to sequence training \cite{  luong15} for identifying the style factors and then add these factors at run time for generating attractive captions. More interestingly, it uses an external monolingual stylized language corpus for training instead of paired images. However, it  uses a new stylized image caption dataset called FlickrStyle10k and can generate captions with different styles.

 Existing image captioning methods consider the factual description about the objects, scene, and their interactions of an image in generating image captions. In our day to day conversations, communications, interpersonal relationships, and decision making we use various stylized and non-factual expressions such as emotions, pride, and shame. However, Mathews et al. \cite {mathews16} claimed that automatic image descriptions are missing this non-factual aspects.  Therefore, they proposed a method called SentiCap. This method can generate image descriptions with positive or negative sentiments. It introduces a novel switching RNN model that combines two CNN+RNNs running in parallel. In each time step, this  switching model generates the probability of switching between two RNNs. One generates captions considering the factual words and other considers the words with sentiments. It then takes inputs from the hidden states of both two RNNs for generating captions. This method can generate captions successfully given the appropriate sentiments.

\subsection{LSTM vs. Others} 

Image captioning intersects computer vision and natural language processing (NLP) research. NLP tasks, in general, can be formulated as a sequence to sequence learning. Several neural language models such as neural probabilistic language model \cite{bengio2003neural}, log-bilinear models \cite{mnih2007three}, skip-gram models \cite{mikolov2013efficient}, and recurrent neural networks (RNNs) \cite{mikolov2010recurrent} have been proposed for learning sequence to sequence tasks. RNNs have widely been used in various sequence learning tasks. However, traditional RNNs suffer from vanishing and exploding gradient problems and cannot adequately handle long-term temporal dependencies.

LSTM \cite{Hochreiter97} networks are a type of RNN that has special units in addition to standard units. LSTM units use a memory cell that can maintain information in memory for long periods of time. In recent years, LSTM based models have dominantly been used in sequence to sequence learning tasks. Another network, Gated Recurrent Unit (GRU) \cite {chung2014empirical} has a similar structure to LSTM but it does not use separate memory cells and uses fewer gates to control the flow of information.

However, LSTMs ignore the underlying hierarchical structure of a sentence. They also require significant storage due to long-term dependencies through a memory cell. In contrast, CNNs can learn the  internal hierarchical structure of the sentences and they are faster in processing than LSTMs. Therefore, recently, convolutional architectures are used in other sequence to sequence tasks, e.g., conditional image generation \cite{van2016conditional} and machine translation \cite{gehring2016convolutional,gehring2017convolutional,vaswani2017attention}.

Inspired by the above success of CNNs in sequence learning tasks, Gu et al. \cite{gu2017empirical} proposed a CNN language model-based image captioning method. This method uses a language-CNN for statistical language modelling. However, the method cannot model the dynamic temporal behaviour of the language model only using a language-CNN. It combines a recurrent network with the language-CNN to model the temporal dependencies properly.
Aneja et al. \cite{aneja2018convolutional} proposed a convolutional architecture for the task of image captioning. They use a feed-forward network without any recurrent function. The architecture of the method has four components: (i) input embedding layer (ii) image embedding layer (iii) convolutional module, and (iv) output embedding layer. It also uses an attention mechanism to leverage spatial image features. They evaluate their architecture on the challenging MSCOCO dataset and shows comparable performance to an LSTM based method on standard metrics. 

Wang et al. \cite{wang2018cnn+} proposed another CNN+CNN based image captioning method. It is similar to the method of Aneja et al. except that it uses a hierarchical attention module to connect the vision-CNN with the language-CNN. The authors of this method also investigate the use of various hyperparameters, including the number of layers and the kernel width of the language-CNN. They show that the influence of the hyperparameters can improve the performance of the method in image captioning.

\section{Datasets and Evaluation Metrics}

 A number of datasets are used for training, testing, and evaluation of the image captioning methods. The datasets differ in various perspective such as the number of images, the number of captions per image, format of the captions, and image size. Three datasets: Flickr8k \cite{hodosh13}, Flickr30k \cite{Plummer15}, and MS COCO Dataset \cite{Lin14} are popularly used. These datasets together with others are described in Section 4.1. In this section, we show sample images with their captions generated by image captioning methods on MS COCO, Flickr30k, and Flickr8k datasets. A number of evaluation metrics are used to measure the quality of the generated captions compared to the ground-truth. Each metric applies its own technique for computation and has distinct advantages. The commonly used evaluation metrics are discussed in Section 4.2. A summary of deep learning-based image captioning methods with their datasets and evaluation metrics are listed in Table 2.

\begin{figure*}
\includegraphics[width=1 \linewidth]{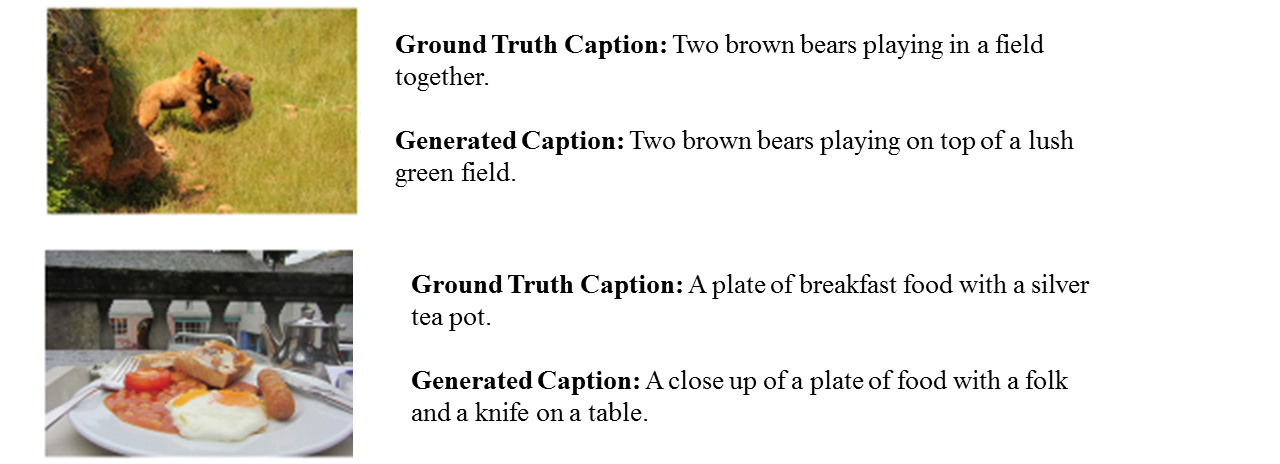}
\caption{Captions generated by Wu et al. \cite{wu15} on some sample images from the MS COCO dataset.}
\end{figure*}

\subsection{ Datasets}

\subsubsection{MS COCO Dataset}

Microsoft COCO Dataset \cite{Lin14} is a very large dataset for image recognition, segmentation, and captioning. There are various features of MS COCO dataset such as object segmentation, recognition in context, multiple objects per class, more than 300,000 images, more than 2 million instances, 80 object categories, and 5 captions per image. Many image captioning methods \cite{Jin15,Wu16E,tran16,wang16,You16, gan17,Pedersoli16, ren17, dai17,shetty17,wu15} use the dataset in their experiments. For example, Wu et al. \cite{wu15} use MS COCO dataset in their method and the generated captions of two sample images are shown in Figure 11. 

\subsubsection{Flickr30K Dataset}

Flickr30K \cite{Plummer15} is a dataset for automatic image description and grounded language understanding. It contains 30k images collected from Flickr with 158k captions provided by human annotators. It does not provide any fixed split of images for training, testing, and validation. Researchers can choose their own choice of numbers for training, testing, and validation. The dataset also contains detectors for common objects, a color classifier, and a bias towards selecting larger objects. Image captioning methods such as \cite{Karpathy15,Vinyals15,wang16,wu2018image,chen2017show} use this dataset for their experiments. For example,  performed their experiment on Flickr30k dataset. The generated captions by Chen et al. \cite{chen2017show} of two sample images of the dataset are shown in Figure 12.

\begin{table*}[]

\centering
\renewcommand{\arraystretch}{0.8}
\begin{tabular}{|l|l|l|}
\hline
\multirow{2}{*}{\textbf{Reference}}     & \multirow{2}{*}{\textbf{Datasets}}                                                    & \multirow{2}{*}{\textbf{Evaluation Metrics}}                                       \\
                               &                                                                              &                                                                           \\ \hline
Kiros et al. 2014~\cite{Kiros14M} & IAPR TC-12,SBU                                                               & BLEU, PPLX                                                                 \\ \hline
Kiros et al. 2014~\cite{Kiros14U}   & Flickr 8K, Flickr 30K                                                        & R@K, mrank                                                                \\ \hline
Mao et al. 2014 ~\cite{Mao14E}     & IAPR TC-12, Flickr 8K/30K                                                    & BLEU, R@K, mrank                                                           \\ \hline
Karpathy et al. 2014~\cite{Karpathy14}    & PASCAL1K, Flickr 8K/30K                                                      & R@K, mrank                                                                \\ \hline
Mao et al. 2015~\cite{Mao14}        & \begin{tabular}[c]{@{}l@{}}IAPR TC-12, Flickr 8K/30K,\\ MS COCO\end{tabular} & BLEU, R@K, mrank                                                           \\ \hline
Chen et al. 2015~\cite{Chen15}        & \begin{tabular}[c]{@{}l@{}}PASCAL, Flickr 8K/30K,\\ MS COCO\end{tabular}     & \begin{tabular}[c]{@{}l@{}}BLEU, METEOR,\\ CIDEr\end{tabular}             \\ \hline
Fang et al. 2015~\cite{Fang15}    & PASCAL, MS COCO                                                              & BLEU, METEOR, PPLX                                                         \\ \hline
Jia et al. 2015~\cite{jia2015}     & Flickr 8K/30K, MS COCO                                                        & BLEU, METEOR, CIDEr                                                        \\ \hline
Karpathy et al. 2015~\cite{Karpathy15}    & Flickr 8K/30K, MS COCO                                                        & BLEU, METEOR, CIDEr                                                        \\ \hline
Vinyals et al. 2015~\cite{Vinyals15}     & Flickr 8K/30K, MS COCO                                                        & BLEU, METEOR, CIDEr                                                        \\ \hline
Xu et al. 2015~\cite{Xu15}         & Flickr 8K/30K, MS COCO                                                        & BLEU, METEOR                                                              \\ \hline
Jin et al. 2015~\cite{Jin15}     & Flickr 8K/30K, MS COCO                                                        & BLEU, METEOR, ROUGE, CIDEr                                                 \\ \hline
Wu et al. 2016~\cite{Wu16E}        & MS COCO                                                                      & BLEU, METEOR, CIDEr                                                        \\ \hline
Sugano et at. 2016~\cite{ sugano16}     & MS COCO                                                                      & BLEU, METEOR, ROUGE, CIDEr                                                 \\ \hline
Mathews et al. 2016~\cite{mathews16} & MS COCO, SentiCap                                                            & BLEU, METEOR, ROUGE, CIDEr                                                 \\ \hline
Wang et al. 2016~\cite{wang16}       & Flickr 8K/30K, MS COCO                                                        & BLEU, R@K                                                                 \\ \hline
Johnson et al. 2016~\cite{ johnson16} & Visual Genome                                                                & METEOR, AP, IoU                                                           \\ \hline
Mao et al. 2016~\cite{ mao16}   & ReferIt                                                                      & BLEU, METEOR, CIDEr                                                        \\ \hline
Wang et al. 2016 \cite{wang2016parallel}       & Flickr 8K                                                        & BLEU, PPL, METEOR                                                                 \\ \hline
Tran et al. 2016~\cite{  tran16}        & \begin{tabular}[c]{@{}l@{}}MS COCO, Adobe-MIT,\\ Instagram\end{tabular}       & Human Evaluation                                                          \\ \hline
Ma et al. 2016 \cite{ma2016describing}       & Flickr 8k, UIUC                                                        & BLEU, R@K                                                                 \\ \hline
You et al. 2016~\cite{You16}        & Flickr 30K, MS COCO                                                           & BLEU, METEOR, ROUGE, CIDEr                                                 \\ \hline
Yang et al. 2016~\cite{yang16}       & Visual Genome                                                                & METEOR, AP, IoU                                                           \\ \hline
Anne et al. 2016~\cite{anne2016deep}        & MS COCO, ImageNet                                                          & BLEU, METEOR                                                                         \\ \hline

Yao et al. 2017~\cite{ Yao16}      & MS COCO                                                                      & BLEU, METEOR, ROUGE, CIDEr                                                 \\ \hline
Lu et al. 2017~\cite{ Lu16}      & Flickr 30K, MS COCO                                                           & BLEU, METEOR, CIDEr                                                        \\ \hline
Chen et al. 2017~\cite{ Chen16}         & Flickr 8K/30K, MS COCO                                                       & BLEU, METEOR, ROUGE, CIDEr                                                 \\ \hline
Gan et al. 2017~\cite{ gan16sem}    & Flickr 30K, MS COCO                                                           & BLEU, METEOR, CIDEr                                                        \\ \hline
Pedersoli et al. 2017~\cite{ Pedersoli16}  & MS COCO                                                                      & BLEU, METEOR, CIDEr                                                        \\ \hline
Ren et al. 2017~\cite{ren17}       & MS COCO                                                                      & BLEU, METEOR, ROUGE, CIDEr                                                 \\ \hline
Park et al. 2017~\cite{Park17}     & Instagram                                                                    & BLEU, METEOR, ROUGE, CIDEr                                                 \\ \hline
Wang et al. 2017~\cite{wang17}   & MS COCO, Stock3M                                                             & SPICE, METEOR, ROUGE, CIDEr                                                \\ \hline
Tavakoli et al. 2017~\cite{ Tavakoli17}   & MS COCO, PASCAL 50S                                                          & BLEU, METEOR, ROUGE, CIDEr                                                 \\ \hline
Liu et al. 2017~\cite{ Liu17}   & Flickr 30K, MS COCO                                                           & BLEU, METEOR                                                              \\ \hline                        
Gan et al. 2017~\cite{  gan17}     & FlickrStyle10K                                                               & BLEU, METEOR, ROUGE, CIDEr                                                 \\ \hline
Dai et al. 2017~\cite{ dai17}      & Flickr 30K, MS COCO                                                           & E-NGAN, E-GAN, SPICE, CIDEr                                                \\ \hline
Shetty et al. 2017~\cite{ shetty17}   & MS COCO                                                                      & \begin{tabular}[c]{@{}l@{}}Human Evaluation,\\ SPICE, METEOR\end{tabular} \\ \hline
Liu et al. 2017~\cite{liu2017improved}    & MS COCO                                                                                  & SPIDEr, Human Evaluation          \\ \hline
Gu et al. 2017~\cite{gu2017empirical} & Flickr 30K, MS COCO                                                                    & BLEU, METEOR, CIDEr, SPICE  \\ \hline
Yao et al. 2017~\cite{yao2017incorporating}    & MS COCO, ImageNet                                                                                & METEOR                 \\ \hline
Rennie et al. 2017~\cite{rennie2017self}    & MS COCO                                                                         &BLEU, METEOR, CIDEr, ROUGE                  \\ \hline
Vsub et al. 2017~\cite{venugopalan2017captioning} & MS COCO, ImageNet                                      & METEOR                                                    \\ \hline

Zhang et al. 2017 ~\cite{zhang2017actor} & MS COCO                                      & BLEU, METEOR, ROUGE, CIDEr                                                    \\ \hline
Wu et al. 2018~\cite{wu2018image}        & Flickr 8K/30K, MS COCO                                                        & BLEU, METEOR, CIDEr                                                        \\ \hline

Aneja et al. 2018 \cite{aneja2018convolutional} & MS COCO                                    & BLEU, METEOR, ROUGE, CIDEr                                                   \\ \hline

Wang et al. 2018 \cite{wang2018cnn+} & MS COCO                                      & BLEU, METEOR, ROUGE, CIDEr                                                     \\ \hline

\end{tabular}
\bigskip
\caption{An overview of methods, datasets, and evaluation metrics}
\label{Table2}
\end{table*}

\begin{figure*}
\includegraphics[width=1 \linewidth]{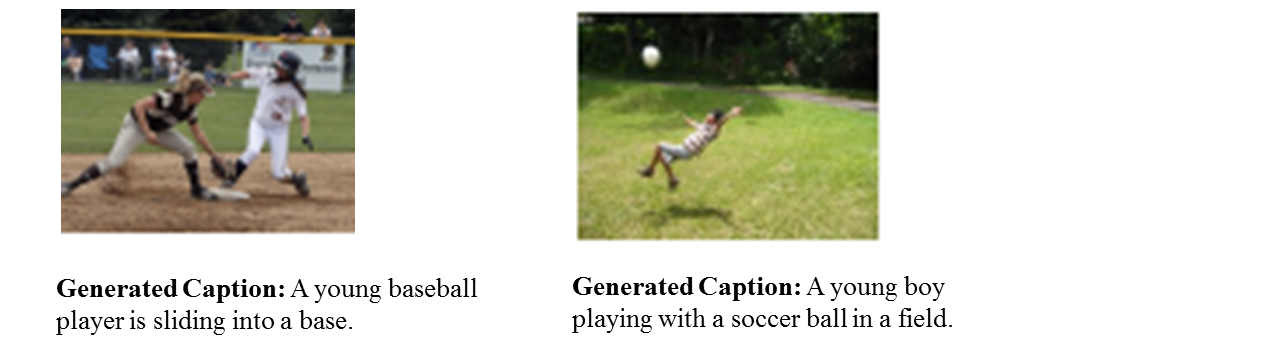}
\caption{Captions generated by Chen et al. \cite{chen2017show} on some sample images from the Flickr30k dataset.}
\end{figure*}

\subsubsection{Flickr8K Dataset}

Flickr8k \cite{hodosh13} is a popular dataset and has 8000 images collected from Flickr. The training data consists of 6000 images, the test and development data, each consists of 1,000 images. Each image in the dataset has 5 reference captions annotated by humans. A number of image captioning methods \cite{jia2015,Jin15,Xu15,wang16,wu2018image,Chen16} have performed experiments using the dataset. Two sample results by Jia et al. \cite{jia2015} on this dataset are shown in Figure 13.

\subsubsection{Visual Genome Dataset}
Visual Genome dataset \cite {krishna17} is another dataset for image captioning. Image captioning requires not only to recognise the objects of an image but it also needs reasoning their interactions and attributes. Unlike the first three datasets where a caption is given to the whole scene, Visual Genome dataset has separate captions for multiple regions in an image. The dataset has seven main parts: region descriptions, objects, attributes, relationships, region graphs, scene graphs, and question answer pairs. The dataset has more than 108k images. Each  image contains an average of 35 objects, 26 attributes, and 21 pairwise relationships between objects.

\subsubsection{Instagram Dataset}

 Tran et al. \cite{tran16} and Park et al. \cite{Park17} created two datasets using images from Instagram which is a photo-sharing social networking services. The dataset of Tran et al. has about 10k images which are mostly from celebrities. However,  Park et al. used their dataset for hashtag prediction and post-generation tasks in social media networks. This dataset contains 1.1m posts on a wide range of topics and a long hashtag lists from 6.3k users.

\begin{figure*}
\includegraphics[width=1 \linewidth]{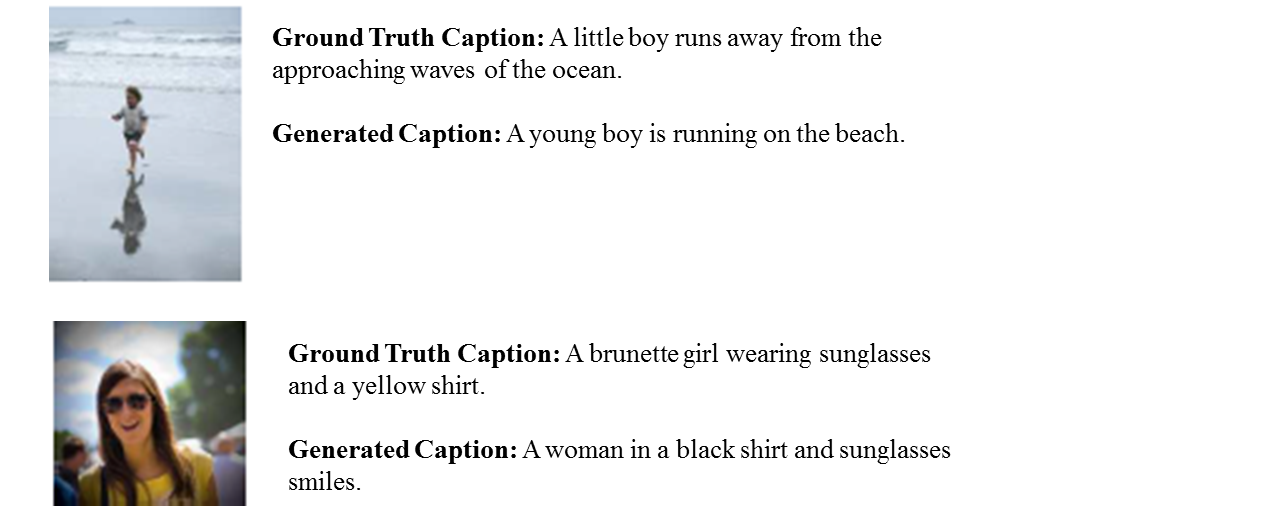}
\caption{Captions generated by Jia et al. \cite{jia2015} on some sample images from the Flickr8k dataset.}
\end{figure*}

\subsubsection{IAPR TC-12 Dataset}

 IAPR TC-12 dataset \cite{grubinger06} has 20k images. The images are collected from various sources such as sports, photographs of people, animals, landscapes and many other locations around the world. The images of this dataset have captions in multiple languages. Images have multiple objects as well.
 
 \subsubsection{Stock3M Dataset}
 Stock3M dataset has 3,217,654 images uploaded by users and it is 26 times larger than MSCOCO dataset. The images of this dataset have a diversity of content.
 
\subsubsection{MIT-Adobe FiveK dataset}
 MIT-Adobe FiveK \cite{bychkovsky11} dataset consists of 5,000 images. These images contain a diverse set of scenes, subjects, and lighting conditions and they are mainly about people, nature, and man-made objects.
 
 \subsubsection{FlickrStyle10k Dataset}
 FlickrStyle10k dataset has 10,000 Flickr images with stylized captions. The training data consists of 7000 images. The validation and test data  consists of 2,000 and 1,000 images respectively. Each image contains romantic, humorous, and factual captions.

\subsection{Evaluation Metrics}

\subsubsection{BLEU}
BLEU (Bilingual evaluation understudy) \cite{papineni02} is a metric that is used to measure the quality of machine generated text. Individual text segments are compared with a set of reference texts and scores are computed for each of them. In estimating the overall quality of the generated text, the computed scores are averaged. However, syntactical correctness is not considered here. The performance of the BLEU metric is varied depending on the number of reference translations and the size of the generated text. Subsequently, Papineni et al. introduced a modified precision metric. This metrics uses n-grams. BLEU is popular because it is a pioneer in automatic evaluation of machine translated text and has a reasonable correlation with human judgements of quality  \cite{denoual05,callison06}. However, it has a few limitations such as BLEU scores are good only if the generated text is short \cite{callison06}. There are some cases  where an increase in BLEU score does not mean that the quality of the generated text is good  \cite{ lin04}.

\subsubsection{ROUGE}

ROUGE (Recall-Oriented Understudy for Gisting Evaluation) \cite{lin2004rouge} is a set of metrics that are used for measuring the quality of text summary. It compares word sequences, word pairs, and n-grams with a set of reference summaries created by humans. Different types of ROUGE such as ROUGE-1, 2, ROUGE-W, ROUGE-SU4 are used for different tasks. For example, ROUGE-1 and ROUGE-W are appropriate for single document evaluation whereas ROUGE-2 and ROUGE-SU4 have good performance in short summaries. However, ROUGE has problems in evaluating multi-document text summary.

\subsubsection{METEOR}
METEOR (Metric for Evaluation of Translation with Explicit ORdering) \cite{banerjee05} is another metric used to evaluate the machine translated language. Standard word segments are compared with the reference texts. In addition to this, stems of a sentence and synonyms of words are also considered for matching. METEOR can make better correlation at the sentence or the segment level. 

\subsubsection{CIDEr}

CIDEr (Consensus-based Image Descripton Evaluation) \cite{vedantam15} is an automatic consensus  metric for evaluating image descriptions. Most existing datasets have only five captions per image. Previous evaluation metrics work with these small number of sentences and are not enough to measure the consensus between generated captions and human judgement. However, CIDEr achieves human consensus using term frequency-inverse document frequency (TF-IDF) \cite{robertson2004}.

\subsubsection{SPICE}

SPICE (Semantic Propositional Image Caption Evaluation) \cite{anderson16} is a new caption evaluation metric based on semantic concept. It is based on a graph-based semantic representation called scene-graph \cite{johnson15,schuster15}. This graph can extract the  information of different objects, attributes and their relationships from the image descriptions.

Existing image captioning methods compute log-likelihood scores to evaluate their generated captions. They use BLEU, METEOR, ROUGE, SPICE, and CIDEr as evaluation metrics. However, BLEU, METEOR, ROUGE are not well correlated with human assessments of quality. SPICE and CIDEr have better correlation but they are hard to optimize. Liu et al. \cite{liu2017improved} introduced a new captions evaluation metric that is a good choice by human raters. It is developed by a combination of SPICE and CIDEr, and termed as SPIDEr. It uses a policy gradient method to optimize the metrics.

The quality of image captioning depends on the assessment of two main aspects: adequacy and fluency. An evaluation metric needs to focus on a diverse set of linguistic features to achieve these aspects. However, commonly used evaluation metrics consider only some specific features (e.g., lexical or semantic) of languages. Sharif et al. \cite{sharif2018learning} proposed learning-based composite metrics for evaluation of image captions. The composite metric incorporates a set of linguistic features to achieve the two main aspects of assessment and shows improved performances.

\section{Comparison on benchmark datasets and common evaluation metrics}

While formal experimental evaluation was left out of the scope of this paper, we present a brief analysis of the experimental results and the performance of various techniques as reported. We cover three sets of results: 
\begin{enumerate}
\item We find a number of methods use the first three datasets listed in Section 4.1. and a number of commonly used evaluation metrics to present the results. These results are shown in Table 3.
\item A few methods fall into the following groups: Attention-based and Other deep learning-based (Reinforcement learning and GAN-based methods) image captioning. The results of such methods are shown in Tables 4 and 5, respectively.
\item We also list the methods that proivide top two results scored on each evaluation metric on the MSCOCO dataset. These results are shown in Table 6.
\end{enumerate}

As shown in Table 3, on Flickr8k, Mao et al. achieved 0.565, 0.386, 0.256, and 0.170 on BLEU-1, BLEU-2, BLEU-3, and BLEU-4 , respectively. For Flickr30k dataset, the scores are 0.600, 0.410, 0.280, and 0.190, respectively which are higher than the Flickr8k scores. The highest scores were achieved on the MSCOCO dataset. The higher results on a larger dataset follows the fact that a large dataset has more data, comprehensive representation of various scenes, complexities, and their own natural context. 
The results of Jia et al. are similar for Flickr8k and Flickr30k datasets but higher on MSCOCO dataset. The method uses visual space for mapping image-features and text features. Mao et al. use multimodal space for the mapping of image-features and text features. On the  other hand, Jia et al. use visual space for the mapping. Moreover, the method uses an Encoder-Decoder architecture where it can guide the decoder part dynamically. Consequently, this method performs better than Mao et al.

Xu et al. also perform better on MSCOCO dataset. This method outperformed both Mao et al. and Jia et al. The main reason behind this is that it uses an attention mechanism which focuses only on relevant objects of the image. The semantic concept-based methods can generate semantically rich captions. Wu et al. proposed a semantic concept-based image captioning method. This method first predicts the attributes of different objects from the image and then adds these attributes with the captions which are semantically meaningful. In terms of performance, the method is superior to all the methods mentioned in Table 3.

\begin{table*}[]
\resizebox{15cm}{!}{
\begin{tabular}{cccccccc}
\hline 
\textbf{Dataset}                    & \textbf{Method}          & \textbf{Category}         & \textbf{BLEU-1} & \textbf{BLEU-2} & \textbf{BLEU-3} & \textbf{BLEU-4} & \textbf{METEOR} \\ \hline 
\multirow{4}{*}{Flickr8k}  & Mao et al. 2015~\cite{Mao14} & MS,SL,WS         & 0.565  & 0.386  & 0.256  & 0.170  & -      \\
                           & Jia et al. 2015~\cite{jia2015} & VS,SL,WS,EDA     & 0.647  & 0.459  & 0.318  & 0.216  & 0.201  \\
                           & Xu et al. 2015~\cite{Xu15}  & VS,SL,WS,EDA,AB  & 0.670  & 0.457  & 0.314  & 0.213  & 0.203  \\
                           & Wu et al. 2018~\cite{ wu2018image}  & VS,SL,WS,EDA,SCB & 0.740  & 0.540  & 0.380  & 0.270  & -      \\ \hline 
\multirow{4}{*}{Flickr30k} & Mao et al. 2015~\cite{Mao14} & MS,SL,WS         & 0.600  & 0.410  & 0.280  & 0.190  & -      \\
                           & Jia et al. 2015~\cite{jia2015} & VS,SL,WS,EDA     & 0.646  & 0.466  & 0.305  & 0.206  & 0.179  \\
                           & Xu et al. 2015~\cite{Xu15} & VS,SL,WS,EDA,AB  & 0.669  & 0.439  & 0.296  & 0.199  & 0.184  \\
                           & Wu et al. 2018~\cite{ wu2018image} & VS,SL,WS,EDA,SCB & 0.730  & 0.550  & 0.400  & 0.280  & -      \\ \hline 
\multirow{4}{*}{MSCOCO}    & Mao et al. 2015~\cite{Mao14} & MS,SL,WS         & 0.670  & 0.490  & 0.350  & 0.250  & -      \\
                           & Jia et al. 2015~\cite{jia2015} & VS,SL,WS,EDA     & 0.670  & 0.491  & 0.358  & 0.264  & 0.227  \\
                           & Xu et al. 2015~\cite{Xu15} & VS,SL,WS,EDA,AB  & 0.718  & 0.504  & 0.357  & 0.250  & 0.230  \\
                           & Wu et al. 2018~\cite{ wu2018image} & VS,SL,WS,EDA,SCB & 0.740  & 0.560  & 0.420  & 0.310  & 0.260 \\ \hline 
\end{tabular}}
\bigskip
\caption{Performance of different image captioning methods on three benchmark datasets and commonly used evaluation metrics.}
\end{table*}

Table 4 shows the results of attention-based based methods on MSCOCO dataset. Xu et al.'s stochastic hard attention produced better results than deterministic soft attention. However, these results were outperformed by Jin et al. which can update its attention based on the scene-specific context.

Wu et al. 2016 and Pedersoli et al. 2017 only show BLEU-4 and METEOR scores which are higher than the aforementioned methods. The method of Wu et al. uses an attention mechanism with a review process. The review process checks the focused attention in every time step and updates it if necessary. This mechanism helps to achieve better results than the prior attention-based methods. Pedersoli et al. propose a different attention mechanism that maps the focused image regions directly with the caption words instead of LSTM state. This behavior of the method drives it to achieve top performances among the mentioned attention-based methods in Table 4.

Reinforcement learning-based (RL) and GAN-based methods are becoming increasingly popular. We name them as ``Other Deep Learning-based Image Captioning". The results of the methods of this group are shown in Table 5. The methods do not have results on commonly used evaluation metrics. However, they have their own potentials to generate the descriptions for the image.

Shetty et al. employed adversarial training in their image captioning method. This method is capable to generate diverse captions. The captions are less-biased with the ground-truth captions compared to the methods use maximum likelihood estimation. To take the advantages of RL, Ren et al. proposed a method that can predict all possible next words for the current word in current time step. This mechanism helps them to generate contextually more accurate captions. Actor-critic of RL are similar to the Generator and the Discriminator of GAN. However, at the beginning of the training, both actor and critic do not have any knowledge about data. Zhang et al. proposed an actor-critic-based image captioning method. This method is capable of predicting the ultimate captions at its early stage and can generate more accurate captions than other reinforcement learning-based methods.

\begin{table*}[]
\resizebox{15cm}{!}{
\begin{tabular}{ccccccccc}
\hline
\multirow{2}{*}{\textbf{Method}} & \multicolumn{1}{l}{\multirow{2}{*}{\textbf{Category}}} & \multicolumn{7}{c}{\textbf{MS COCO}}                                                                                                                                                                              \\ \cline{3-9}
                        & \multicolumn{1}{l}{}                          & \multicolumn{1}{l}{\textbf{BLEU-1}} & \multicolumn{1}{l}{\textbf{BLEU-2}} & \multicolumn{1}{l}{\textbf{BLEU-3}} & \multicolumn{1}{l}{\textbf{BLEU-4}} & \multicolumn{1}{l}{\textbf{METEOR}} & \multicolumn{1}{l}{\textbf{ROUGE-L}} & \multicolumn{1}{l}{\textbf{CIDEr}} \\ \hline
Xu et al. 2015~\cite{Xu15}, soft    & VS,SL,WS,EDA,VC                               & 0.707                      & 0.492                      & 0.344                      & 0.243                      & 0.239                      & -                           & -                         \\
Xu et al. 2015~\cite{Xu15}, hard    & VS,SL,WS,EDA,VC                               & 0.718                      & 0.504                      & 0.357                      & 0.250                      & 0.230                      & -                           & -                         \\
Jin et al. 2015~\cite{Jin15}         & VS,SL,WS,EDA,VC                               & 0.697                      & 0.519                      & 0.381                      & 0.282                      & 0.235                      & 0.509                       & 0.838                     \\
Wu et al. 2016~\cite{Wu16E}         & VS,SL,WS,EDA,VC                               & -                          & -                          & -                          & 0.290                      & 0.237                      & -                           & 0.886                     \\
Pedersoli et al. 2017~\cite{ Pedersoli16}  & VS,SL,WS,EDA,VC                               & -                          & -                          & -                          & 0.307                      & 0.245                      & -                           & 0.938      \\ \hline              
\end{tabular}}
\bigskip
\caption{Performance of attention-based image captioning methods on MSCOCO dataset and commonly used evaluation metrics.}
\end{table*}

\fancypagestyle{alim}{\fancyhf{}\renewcommand{\headrulewidth}{0pt}\fancyfoot[L]{\textit{*A dash (-) in the tables of this paper indicates results are unavailable}}}
\thispagestyle{alim}

\begin{table*}[]
\resizebox{15cm}{!}{
\begin{tabular}{cccccccccc}
\hline
\multirow{2}{*}{\textbf{Method}} & \multirow{2}{*}{\textbf{Category}} & \multicolumn{8}{c}{\textbf{MS COCO}}                                         \\ \cline{3-10}
                        &                           & \textbf{BLEU-1} & \textbf{BLEU-2} & \textbf{BLEU-3} & \textbf{BLEU-4} & \textbf{METEOR} & \textbf{ROUGE-L} & \textbf{CIDEr} & \textbf{SPICE} \\ \hline
Shetty et al. 2017\textsubscript{GAN}~\cite{shetty17}      & VS,ODL,WS,EDA             & -      & -      & -      & -      & 0.239  & -       & -     & 0.167 \\
Ren et al. 2017\textsubscript{RL}~\cite{ren17}         & VS,ODL,WS,EDA             & 0.713  & 0.539  & 0.403  & 0.304  & 0.251  & 0.525   & 0.937 & -     \\
Zhang et al. 2017\textsubscript{RL}~\cite{zhang2017actor}       & VS,ODL,WS,EDA             & -      & -      & -      & 0.344  & 0.267  & 0.558   & 1.162 & -    \\ \hline
\end{tabular}}
\bigskip
\caption{Performance of Other Deep learning-based image captioning methods on MSCOCO dataset and commonly used evaluation metrics.}
\end{table*}

\begin{table*}[]
\resizebox{15cm}{!}{
\begin{tabular}{cccccccccc}
\hline
\multirow{2}{*}{\textbf{Method}} & \multirow{2}{*}{\textbf{Category}} & \multicolumn{8}{c}{\textbf{MSCOCO}}                                                                                                                                                                                   \\ \cline{3-10}
                        &                           & \textbf{BLEU-1}                  & \textbf{BLEU-2}                  & \textbf{BLEU-3}                  & \textbf{BLEU-4}                  & \textbf{METEOR}                  & \textbf{ROUGE-L}                 & \textbf{CIDEr}                   & \textbf{SPICE}                   \\ \hline
Lu et al. 2017~\cite{Lu16}          & VS,SL,WS,EDA,AB           & \textit{\textbf{0.742}} & \textit{\textbf{0.580}} & \textbf{0.439}          & 0.332                   & \textbf{0.266}          & -                       & \textbf{1.085}          & -                       \\
Gan et al. 2017~\cite{gan16sem}     & VS,SL,WS,CA,SCB           & \textbf{0.741}          & \textbf{0.578}          & \textit{\textbf{0.444}} & \textbf{0.341}          & 0.261                   & -                       & 1.041                   & -                       \\
Zhang et al. 2017~\cite{zhang2017actor}       & VS,ODL,WS,EDA             & -                       & -                       & -                       & \textit{\textbf{0.344}} & \textit{\textbf{0.267}} & \textit{\textbf{0.558}} & \textit{\textbf{1.162}} & -                       \\
Rennie et al. 2017~\cite{rennie2017self}      & VS,ODL,WS,EDA             & -                       & -                       & -                       & .319                    & 0.255                   & \textbf{0.543}          & 1.06                    & -                       \\
Yao et al. 2017~\cite{Yao16}      & VS,SL,WS,EDA,SCB          & 0.734                   & 0.567                   & 0.430                   & 0.326                   & 0.254                   & 0.540                   & 1.00                    & \textit{\textbf{0.186}} \\
Gu et al. 2017~\cite{gu2017empirical}          & VS,SL,WS,EDA              & 0.720                   & 0.550                   & 0.410                   & 0.300                   & 0.240                   & -                       & 0.960                   & \textbf{0.176}     \\ \hline    
\end{tabular}}
\bigskip
\caption{Top two methods based on different evaluation metrics and MSCOCO dataset (Bold and Italic indicates the best result; Bold indicates the second best result).}
\end{table*}

We found that the performance of a technique can vary across different metrics. Table 6 shows the methods based on the top two scores on every individual evaluation metric. For example, Lu et al., Gan et al., and Zhang et al. are within the top two methods based on the scores achieved on BLEU-n and METEOR metrics. BLEU-n metrics use variable length phrases of generated captions to match against ground-truth captions. METEOR \cite{banerjee05} considers the precision, recall, and the alignments of the matched tokens. Therefore, the generated captions by these methods have good precision and recall accuracy as well as the good similarity in word level. ROUGE-L evaluates the adequacy and fluency of generated captions, whereas CIDEr focuses on grammaticality and saliency. SPICE can analyse the semantics of the generated captions. Zhang et al., Rennie et al., and Lu et al. can generate captions, which have adequacy, fluency, saliency, and are grammaticality correct than other methods in Table 6. Gu et al. and Yao et al. perform well in generating semantically correct captions.

\section{Discussions and Future Research Directions}

Many deep learning-based methods have been proposed for generating automatic image captions in the recent years. Supervised learning, reinforcement learning, and GAN based methods are commonly used in generating image captions. Both visual space and multimodal space can be used in supervised learning-based methods. The main difference between visual space and multimodal space occurs in mapping. Visual space-based methods perform explicit mapping from images to descriptions. In contrast, multimodal space-based methods incorporate implicit vision and  language models. Supervised learning-based methods are further categorized into Encoder-Decoder architecture-based, Compositional architecture-based, Attention-based, Semantic concept-based, Stylized captions, Dense image captioning, and Novel object-based image captioning.

 Encoder-Decoder architecture-based methods use a simple CNN and a text generator for generating image captions. Attention-based image captioning methods focus on different salient parts of the image and achieve better performance than encoder-decoder architecture-based methods. Semantic concept-based image captioning methods selectively focus on different parts of the image and can generate semantically rich captions. Dense image captioning methods can generate region based image captions. Stylized image captions express various emotions such as romance, pride, and shame. GAN and RL based image captioning methods can generate diverse and multiple captions.

 MSCOCO, Flickr30k and Flickr8k dataset are common and popular datasets used for image captioning. MSCOCO dataset is very large dataset and all the images in these datasets have multiple captions. Visual Genome dataset is mainly used for region based image captioning. Different evaluation metrics are used for measuring the performances of image captions. BLEU metric is good for small sentence evaluation. ROUGE has different types and they can be used for evaluating different types of texts. METEOR can perform an evaluation on various segments of a caption. SPICE is better in understanding semantic details of captions compared to other evaluation metrics.

Although success has been achieved in recent years, there is still a large scope for improvement. Generation based methods can generate novel captions for every image. However, these methods fail to detect prominent objects and attributes and their relationships to some extent in generating accurate and multiple captions. In addition to this, the accuracy of the generated captions largely depends on syntactically correct and diverse captions which in turn rely on powerful and sophisticated language generation model. Existing methods show their performances on the datasets where images are collected from the same domain. Therefore, working on open domain dataset will be an interesting avenue for research in this area. Image-based factual descriptions are not enough to generate high-quality captions. External knowledge can be added in order to generate attractive image captions. Supervised learning needs a large amount of labelled data for training. Therefore, unsupervised learning and reinforcement learning will be more popular in future in image captioning.

\section{Conclusions} 
In this paper, we have reviewed deep learning-based image captioning methods. We have given a taxonomy of image captioning techniques, shown generic block diagram of the major groups and highlighted their pros and cons. We discussed different evaluation metrics and datasets with their strengths and weaknesses. A brief summary of experimental results is also given. We briefly outlined potential research directions in this area. Although deep learning-based image captioning methods have achieved a remarkable progress in recent years, a robust image captioning method that is able to generate high quality captions for nearly all images is yet to be achieved. With the advent of novel deep learning network architectures, automatic image captioning will remain an active research area for some time.

\section*{Acknowledgements}

This work was partially supported by an Australian Research Council grant DE120102960.

\bibliographystyle{ACM-Reference-Format}
\bibliography{sample-acmsmall}

\end{document}